\documentclass{article}

\PassOptionsToPackage{numbers, compress}{natbib}
\usepackage[final]{neurips_2022}

\usepackage[utf8]{inputenc} %
\usepackage[T1]{fontenc}    %
\usepackage{hyperref}       %
\usepackage{url}            %
\usepackage{booktabs}       %
\usepackage{amsfonts}       %
\usepackage{nicefrac}       %
\usepackage{microtype}      %
\usepackage{wrapfig}
\usepackage{xcolor}         %
\usepackage{caption}
\usepackage{subcaption}
\usepackage{inconsolata}
\usepackage{bbm}
\usepackage{algpseudocode}
\usepackage{amsmath}
\usepackage{amssymb}
\usepackage{graphicx}

\usepackage{verbatim}
\makeatletter
\def\verbatim@font{\scriptsize\ttfamily}
\makeatother

\usepackage{booktabs} %

\usepackage{hyperref}

\usepackage{algorithm}
\usepackage{algorithmicx}

\title{Improving Intrinsic Exploration with\\Language Abstractions}

\author{%
    Jesse Mu\textsuperscript{1}\thanks{Work done while at Meta AI. Correspondence to \texttt{muj@stanford.edu}}
    , Victor Zhong\textsuperscript{2,3}, Roberta Raileanu\textsuperscript{3}, Minqi Jiang\textsuperscript{3,4},\\\textbf{Noah Goodman\textsuperscript{1}, Tim Rockt\"{a}schel\textsuperscript{4*}, Edward Grefenstette\textsuperscript{4,5*}}
    \\
    \textsuperscript{1}Stanford University, \textsuperscript{2}University of Washington, \textsuperscript{3}Meta AI, \textsuperscript{4}University College London,
    \textsuperscript{5}Cohere
}

\begin{document}

\maketitle

\begin{abstract}
Reinforcement learning (RL) agents are particularly hard to train when rewards are sparse. One common solution is to use \emph{intrinsic} rewards to encourage agents to explore their environment. However, recent intrinsic exploration methods often use state-based novelty measures which reward low-level exploration and may not scale to domains requiring more abstract skills. Instead, we explore \emph{language} as a general medium for highlighting relevant abstractions in an environment. Unlike previous work, we evaluate whether language can improve over existing exploration methods by directly extending (and comparing to) competitive intrinsic exploration baselines: AMIGo (Campero et al., 2021) and NovelD (Zhang et al., 2021).
These language-based variants outperform their non-linguistic forms by 47--85\% across 13 challenging tasks from the MiniGrid and MiniHack environment suites.
\end{abstract}

\section{Introduction}
\label{sec:introduction}

A central challenge in reinforcement learning (RL) is designing agents that can solve complex, long-horizon tasks with sparse rewards. In the absence of extrinsic rewards, one popular solution is to provide \emph{intrinsic} rewards for exploration \cite{oudeyer2009intrinsic,oudeyer2007intrinsic,schmidhuber1991possibility,schmidhuber2010formal}. This invariably leads to the challenging question: how should one measure exploration?
One common answer is that an agent should be rewarded for attaining ``novel'' states in the environment, but naive measures of novelty have limitations.
For example, consider an agent that starts in the kitchen of a large house and must make an omelet. Simple state-based exploration will reward an agent for visiting every room in the house, but a more effective strategy would be to stay put and use the stove. Moreover, like kitchens with different-colored appliances, states can look cosmetically different but have the same underlying semantics, and thus are not truly novel. Together, these constitute two fundamental challenges for intrinsic exploration: first, how can we reward true progress in the environment over meaningless exploration? Second, how can we tell when a state is not just superficially, but \emph{semantically} novel?

Fortunately, humans are equipped with a powerful tool for solving both problems: language. As a cornerstone of human intelligence, language has strong priors over the features and behaviors needed for exploration and skill acquisition. It also describes a rich and compositional set of meaningful behaviors as simple as directions (e.g.\ \emph{move left}) and as abstract as conjunctions of high level tasks (e.g.\ \emph{retrieve the ring and defeat the wizard}) that can categorize and unify many possible world states.

\begin{wrapfigure}{r}{0.5\textwidth}
    \centering
    \vspace{-0.75em}
    \includegraphics[width=0.48\textwidth]{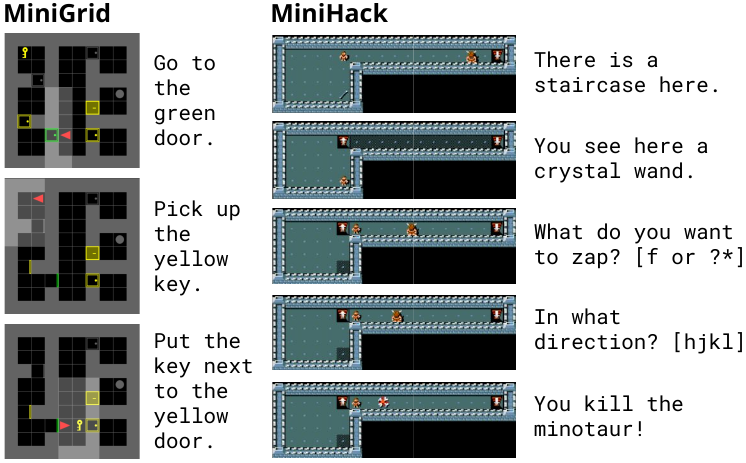}
    \caption{\textbf{Language conveys meaningful environment abstractions.} Language  state annotations in the MiniGrid KeyCorridorS4R3 \cite{chevalierboisvert2018minimalistic} and MiniHack Wand of Death (Hard) \cite{samvelyan2021minihack} tasks.}
    \vspace{-3em}
    \label{fig:language_overview}
\end{wrapfigure}

Our aim is to see whether language abstractions can improve existing state-based exploration methods in RL.
While language-guided exploration methods exist in the literature \cite{bahdanau2018learning,blukis2019learning,colas2020languageconditioned,colas2020language,fu2019language,goyal2019using,goyal2020pixl2r,harrison2018guiding,mirchandani2021ella,schwartz2019language,tasrin2021influencing,waytowich2019narration}, we make two key contributions over prior work. First, existing methods assume access to a high-level linguistic instruction for reward shaping, or otherwise assume that any intermediate language annotations encountered are always helpful for learning.
Instead, we study settings without instructions, with more diverse intermediate messages (Figure~\ref{fig:language_overview}) that may or may not be useful, but may nonetheless be a more effective measure of novelty than raw states.

Second, past work often compares only to vanilla RL, while ignoring competitive intrinsic exploration baselines. This leaves the true utility of language over simpler state-based exploration unclear.
To remedy this issue, we conduct a controlled evaluation on the effect of language on competitive approaches to exploration by extending two recent, state-of-the-art methods:
AMIGo \cite{campero2021learning}, where a teacher proposes intermediate location-based goals for a student, and NovelD \cite{zhang2021noveld}, which rewards an agent for visiting novel regions of the state space. Building upon these methods, we propose \textbf{L-AMIGo}, where the teacher proposes goals expressed via language instead of coordinates, and \textbf{L-NovelD}, a variant of NovelD with an additional exploration bonus for visiting linguistically-novel states.
Across 13 challenging, procedurally-generated, sparse-reward tasks in the MiniGrid \cite{chevalierboisvert2018minimalistic} and MiniHack \cite{samvelyan2021minihack} environment suites, we show that language-parameterized exploration methods outperform their non-linguistic counterparts by 47--85\%, especially in more abstract tasks with larger state and action spaces. We also show that language improves the interpretability of the training process, either by developing a natural curriculum of semantic goals (in L-AMIGo) or by allowing us to visualize the most novel language during training (in L-NovelD). Finally, we show when and where the fine-grained compositional semantics of the language improves agent exploration, when compared to non-compositional baselines.

\section{Related Work}
\label{sec:related-work}

\paragraph{Exploration in RL.}
Exploration has a long history in RL, from $\varepsilon$-greedy \cite{sutton1998reinforcement} or count-based exploration \cite{bellemare2016unifying,machado2020count,Martin2017CountBasedEI,ostrovski2017count,strehl2008analysis,tang2017exploration} to intrinsic motivation \cite{oudeyer2008can,oudeyer2009intrinsic,oudeyer2007intrinsic} and curiosity-based learning \cite{schmidhuber1991possibility}. More recently, deep neural networks have been used to measure novelty with changes in state representations \cite{burda2018exploration,raileanu2020ride,zhang2021noveld} or prediction errors in world models \cite{achiam2017surprise,pathak2017curiosity,stadie2015incentivizing}. Another long tradition generates curricula of intrinsic goals to encourage learning \cite{campero2021learning,colas2020language,colas2020autotelic,Fang2019CurriculumguidedHE,florensa2017reverse,forestier2017intrinsically,portelas2020teacher,portelas2020automatic,racaniere2020automated}. In this paper, we explore the potential benefit of language on these approaches to exploration.

\paragraph{Language for Exploration.}

The observation that language-guided exploration can improve RL is not new: language has been used to shape policies \cite{harrison2018guiding,tasrin2021influencing} and rewards \cite{bahdanau2018learning,blukis2019learning,fu2019language,goyal2019using,goyal2020pixl2r,mirchandani2021ella,schwartz2019language,waytowich2019narration} and set intrinsic goals \cite{colas2020languageconditioned,colas2020language}.
Crucially, our work differs from prior work in two ways: first, instead of reward shaping with high-level instructions, we use noisier, intermediate language annotations for exploration; second, we directly extend and compare to competitive intrinsic exploration baselines.

L-AMIGo, our variant of AMIGo with language goals, is similar to the IMAGINE agent of \citet{colas2020language}, which also sets intrinsic language goals.
However, IMAGINE is built for instruction following, and requires a perfectly compositional space of language goals, which the agent tries to explore so that it can complete novel goals at test time. Instead, we make no assumptions on the language and explore to maximize extrinsic reward, using an alternative \emph{goal difficulty} metric to measure progress.

Meanwhile, reward shaping and inverse RL methods \cite{bahdanau2018learning,blukis2019learning,fu2019language,goyal2019using,goyal2020pixl2r,harrison2018guiding,mirchandani2021ella,schwartz2019language,tasrin2021influencing,waytowich2019narration} reward an agent for actions associated with linguistic descriptions, but again are primarily designed for instruction following, where an extrinsic goal is available to help shape intermediate rewards. In our setting, however, we have not high-level extrinsic goals but low-level \emph{intermediate} language annotations. Extrinsic reward shaping methods such as LEARN \cite{goyal2019using} could be naively applied by simply doing reward shaping with every intermediate language annotation, and a few of these methods are designed for low-level language subgoals \cite{harrison2018guiding,mirchandani2021ella}. However, a shared assumption of these approaches is that \emph{language is always helpful}: either because we have expert-curated messages (as in \citet{harrison2018guiding}), or because we have goal descriptions that let us identify subgoals relevant to the extrinsic goal (as in ELLA; \citet{mirchandani2021ella}). In our tasks, however, most language is \emph{unhelpful} for progress in the environment, and we have no extrinsic goals. Consequently, past methods reduce to simply giving a fixed reward for every intermediate message encountered, which (we will show) fails to learn. Finally, work concurrent to ours by \citet{tam2022semantic} tackles similar ideas in photorealistic environments that permit transfer from foundation models. Instead, we explore domain-specific symbolic games with built-in language where such models are not readily available.

A final distinguishing contribution of our work is that prior work often neglects non-linguistic exploration baselines. For example, Harrison et al. \cite{harrison2018guiding} and LEARN \cite{goyal2019using} compare to vanilla RL only; ELLA \cite{mirchandani2021ella} compares to LEARN and RIDE \cite{raileanu2020ride}, with limited improvements over RIDE. Prior work can thus be summarized as showing that linguistic rewards improve over extrinsic rewards alone. Instead, we provide novel evidence that \emph{linguistic rewards improve upon state-based intrinsic rewards}, using the same exploration methods and challenging tasks typical of recent work in RL.

\section{Problem Statement}
\label{sec:problem_statement}
\begin{figure}[t]
    \captionsetup{justification=raggedright}
    \captionsetup{width=0.28\linewidth}
    {
    \centering
    \includegraphics[width=\textwidth]{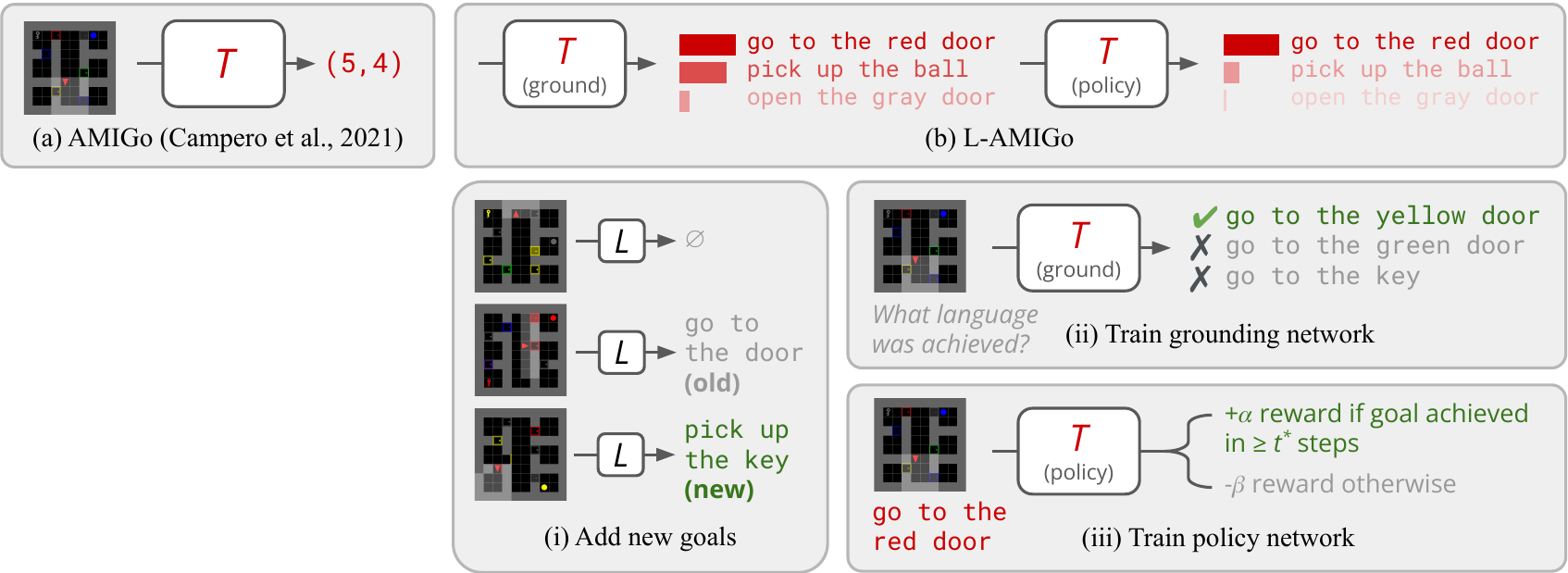}
    }
    \begin{minipage}[t]{1.39in}
    \vspace{-1.5in}
    \caption{\textbf{L-AMIGo}. (a) AMIGo. (b) L-AMIGo teacher first predicts achievable goals, then samples a goal. (i--iii) L-AMIGo teacher training steps: updating the goal set $\mathcal{G}$ and training grounding and policy networks.}
    \label{fig:lamigo_overview}
    \end{minipage}
    \vspace{-1em}
\end{figure}
We explore RL in the setting of an augmented Markov Decision Process (MDP) defined as the tuple ($\mathcal{S}, \mathcal{A}, T, R, \gamma, L$), where $\mathcal{S}$ and $\mathcal{A}$ are the state and action spaces, $T : \mathcal{S} \times \mathcal{A} \rightarrow \mathcal{S}$ is the environment transition dynamics, $R : \mathcal{S} \times \mathcal{A} \rightarrow \mathbb{R}$ is the \emph{extrinsic} reward function, where $r_t = R(s_{t}, a_{t})$ is the reward obtained at time $t$ by taking action $a_t$ in state $s_t$, and $\gamma$ is the discount factor.
To add language, we assume access to an \emph{annotator} $L$ that produces \textbf{language descriptions} for states: $\ell_t = L(s_t)$, such as those in Figure~\ref{fig:language_overview}. Note that not every state needs a description (which we model with a null description $\varnothing$) and the set of descriptions need not be known ahead of time.\footnote{For presentational simplicity, the annotator here outputs a single description per state, but in practice, we allow an annotator to produce \emph{multiple} descriptions: e.g.\ in MiniGrid, \emph{open the door} and \emph{open the red door} describe the same state. This requires two minor changes in the equations, described in Footnotes~\ref{foot:grounding_multi_change} and~\ref{foot:noveld_multi_change}.}
We ultimately seek a policy that maximizes the expected discounted (extrinsic) reward $R_t = \mathbb{E}[ \sum_{k = 0}^H \gamma^k r_{t + k} ]$, where $H$ is the finite time horizon. During training, however, we maximize an \emph{augmented} reward $r^+_t = r_t + \lambda r^i_t$, where $r^i_t$ is an \emph{intrinsic} reward  and $\lambda$ is a scaling hyperparameter.

Like past work \cite{jiang2019language,mirchandani2021ella,waytowich2019narration} we make the simplifying assumption of access to an oracle language annotator $L$ provided by the environment. Note that the language annotator is ``oracle'' in that it always outputs messages that are true of the current state, but \emph{not} ``oracle'' in that it indiscriminately outputs messages that are not necessarily relevant for the extrinsic goal. Many modern RL environments are pre-equipped with language, including NetHack/MiniHack \cite{kuttler2020nethack,samvelyan2021minihack}, text-based games \cite{cote2018textworld,shridhar2021alfworld,urbanek2019learning}, and in fact most video games in general. In the absence of an oracle annotator, one common approach is to learn an annotator model from a dataset of language-annotated states \cite{bahdanau2018learning,goyal2019using,mirchandani2021ella}, though such datasets are often generated from oracles that are simply run offline instead \cite{mirchandani2021ella}. Concurrent work \cite{tam2022semantic} uses pretrained foundation models to automatically provide annotations in 3D environments, though such models are not readily available in the symbolic 2D games we explore. Since this idea has been well-proven, we assume oracle access to $L$, but as an example, one could straightforwardly adapt the annotator model trained on BabyAI by \citet{mirchandani2021ella} to our setting.

\section{L-AMIGo}

We now describe our approach to jointly training a student and a goal-proposing teacher, extending AMIGo \cite{campero2021learning} to arbitrary language goals.

\subsection{Adversarially Motivated Intrinsic Goals (AMIGo)}

AMIGo \cite{campero2021learning} augments an RL student policy with goals generated by a teacher, which provide intrinsic reward when completed (Figure~\ref{fig:lamigo_overview}a). The idea is that the teacher should propose intermediate goals that start simple, but grow harder to encourage an agent to explore its environment.

\paragraph{Student.} Formally, the student $S$ is a \emph{goal-conditioned} policy parameterized as $\pi_S(a_t \mid s_t, g_t; \theta_S)$, where $g_t$ is the goal provided by the teacher, and the student receives an intrinsic reward $r^i_t$ of 1 only if the teacher's goal at that timestep is completed. The student receives a goal from the teacher either at the beginning of an episode, or mid-episode, if the previous goal has been completed.

\paragraph{Teacher.} Separately, AMIGo trains an adversarial \emph{teacher} policy $\pi_T(g_t \mid s_0; \theta_T)$ to propose goals to the student given its initial state.
The teacher is trained with a reward $r^T_t$ that depends on a \emph{difficulty threshold} $t^*$: the teacher is given a positive reward of $+\alpha$ for proposing goals that take the student more than $t^*$ timesteps to complete, and $-\beta$ for goals that are completed sooner, or never completed within the finite time horizon. To encourage proposing harder and harder goals that promote exploration, $t^*$ is increased linearly throughout training: whenever the student completes 10 goals in a row under the current difficulty threshold, it is increased by 1, up to some tunable maximum difficulty. Finally, to encourage intermediate goals that are aligned with the extrinsic goal, the teacher is also rewarded with the extrinsic reward when the student attains it.

This teacher is updated separately from the student at different time intervals. Formally, its training data is batches of $(s_0, g_t, r^T_t)$ tuples collected from student trajectories for nonzero $r^T_t$, where $s_0$ is the initial state of the student's trajectory and $g_t$ is the goal that led to reward $r^T_t$.

The original paper \cite{campero2021learning} implements AMIGo for MiniGrid only, where the goals $g_t$ are $(x, y)$ coordinates to be reached. The student gets the goal embedded directly in the $M \times N$ environment, and the teacher is a dimensionality-preserving convolutional network which encodes the student's $M \times N$ environment into an $M \times N$ distribution over coordinates, from which a single goal is selected.

\subsection{Extension to L-AMIGo}

\textbf{Student.} The L-AMIGo student is a policy conditioned not on $(x, y)$ goals, but on \emph{language goals} $\ell_t$: $\pi_{S}(a_t \mid s_t, \ell_t; \theta_S)$. Given the ``goal'' $\ell_t$, the student is now rewarded if it reaches a state with the language description $\ell_t$, i.e.\ if $\ell_t = L(s_t)$.\footnote{We can treat language \emph{goals} and \emph{state descriptions} equivalently, even if the wordings are slightly different across environments. In MiniGrid, messages (e.g.\ \emph{go to the red door}) look like goals but can also be interpreted as state descriptions: \emph{[in this state, you have] go[ne] to the red door} In MiniHack, messages are description-like (e.g.\ \emph{you kill the minotaur!}), but imagine the teacher's goal as \emph{[reach a state where] you kill the minotaur!}}. Typically this student will encode the goal with a learned language model and concatenate the goal representation with its state representation.

\paragraph{Teacher.} Now the L-AMIGo teacher selects goals from the set of possible language descriptions in the environment. Because the possible goals are initially unknown, the teacher maintains a running set of goals $\mathcal{G}$ that is updated as the student encounters new state descriptions (Figure~\ref{fig:lamigo_overview}i).

This move to language creates a challenge: not only must a teacher choose a goal to propose, it must also determine which goals are achievable at all. For example, the goal \emph{go to the red door} only makes sense in environments with red doors. In L-AMIGo, these tasks are factorized into a \textbf{policy network}, which produces the distribution over goals given a student's state, and a \textbf{grounding network}, which predicts the probability that a goal is likely to be achieved in the first place (Figure~\ref{fig:lamigo_overview}b):
\begin{align}
    \pi_T(\ell_t \mid s_t; \theta_T) &\propto p_{\text{ground}}(\ell_t \mid s_t; \theta_T) \cdot p_{\text{policy}}(\ell_t \mid s_t; \theta_T) \\
    p_\text{ground}(\ell_t \mid s_t; \theta_T) &= \sigma\left( f(\ell_t; \theta_T) \cdot h_{\text{ground}}(s_t; \theta_T) \right) \label{eq:lamigogrounding} \\
    p_{\text{policy}}(\ell_t \mid s_t; \theta_T) &\propto f(\ell_t; \theta_T) \cdot h_\text{policy}(s_t; \theta_T) \label{eq:lamigopolicy}
\end{align}
Equation~\ref{eq:lamigopolicy} describes the policy network as producing a probability for a goal by computing the dot product between goal and state representations $f(\ell_t; \theta_T)$ and $h_\text{policy}(s_t; \theta_T)$, normalizing over possible goals; this policy is learned identically to the standard AMIGo teacher (Figure~\ref{fig:lamigo_overview}iii).
Equation~\ref{eq:lamigogrounding} specifies the grounding network as predicting whether a goal is \emph{achievable} in an environment, by applying the sigmoid function to the dot product between the goal representation $f(\ell_t; \theta_T)$ and a (possibly separate) state representation $h_\text{ground}(s_t; \theta_T)$. Given an oracle grounding classifier, which outputs only 0 or 1, this is equivalent to restricting the teacher to proposing only goals that are achievable in a given environment. In practice, however, we learn the classifier online (Figure~\ref{fig:lamigo_overview}ii). Given the initial state $s_0$ of an episode, we ask the grounding network to predict the first language description encountered along this trajectory: $\ell_{\text{1st}} = L(s_{t'})$, where $t'$ is the minimum $t$ where $L(s_t) \neq \varnothing$. This is formalized as a multilabel binary cross entropy loss,
\begin{align}
    \mathcal{L}_{\text{ground}}(s_0, \ell_\text{1st}) = - \log (p_{\text{ground}}( \ell_{\text{1st}} \mid s_0 ; \theta_T)) - \tfrac{1}{|\mathcal{G}| - 1} \sum_{\ell' \in \mathcal{G} \setminus \{ \ell_{\text{1st}} \}} \log (1 - p_\text{ground}(\ell' \mid s_0 ; \theta_T) ), \label{eq:lamigogroundingxent}
\end{align}
where the second term noisily generates negative samples of (start state, unachieved description) pairs based on the set of descriptions $\mathcal{G}$ known to the teacher at the time, similar to contrastive learning.\footnote{If multiple ``first'' descriptions are found, the teacher predicts 1 for \emph{each} description, and 0 for all others.\label{foot:grounding_multi_change}}
Note that since $\mathcal{G}$ is updated during training, Equation~\ref{eq:lamigogroundingxent} grows to include more terms over time.

To summarize, training the teacher involves three steps: (1) updating the running set of descriptions seen in the environment, (2) learning the policy network based on whether the student achieved goals proposed by the teacher, and (3) learning the grounding network by predicting descriptions encountered from initial states. Algorithm~\ref{alg:lamigo} in Appendix~\ref{app:lamigo_algorithm} describes how L-AMIGo trains in an asynchronous actor-critic framework, where the student and teacher are jointly trained from batches of experience collected from separate actor threads, as used in our experiments (see Section~\ref{sec:experiments}).

\section{L-NovelD}

Next, we describe NovelD \cite{zhang2021noveld},
which extends simpler tabular- \cite{strehl2008analysis} or pseudo- \cite{bellemare2016unifying,burda2018exploration} count-based intrinsic exploration methods,
and our language variant, L-NovelD.
Instead of simply rewarding an agent for rare states, NovelD rewards agents for \emph{transitioning} from states with low novelty to states with higher novelty. \citet{zhang2021noveld} show that NovelD surpasses Random Network Distillation \cite{burda2018exploration}, another popular exploration method, on a variety of tasks including MiniGrid and Atari.

\subsection{NovelD}

NovelD defines the reward $r^i_t$ to be the difference in novelty between state $s_t$ and previous state $s_{t - 1}$:
\begin{align}
    r^i_t =\; &\text{NovelD}_s(s_t, s_{t - 1}) \triangleq \underbrace{\max(N(s_{t}) - \alpha N(s_{t - 1}), 0)}_{\text{Term 1 (NovelD)}} \cdot  
    \underbrace{\mathbbm{1}(N_e(s_t) = 1)}_{\text{Term 2 (ERIR)}}. \label{eq:noveld}
\end{align}
In the first NovelD term, $N(s_t)$ is the novelty of state $s_t$; this quantity describes the difference in novelty between successive states, which is clipped $> 0$ so the agent is not penalized from moving back to less novel states. $\alpha$ is a hyperparameter that scales the average magnitude of the reward.
The second term is the \textbf{E}pisodic \textbf{R}eduction on \textbf{I}ntrinsic \textbf{R}eward (ERIR): a constraint that the agent only receives reward when encountering a state \emph{for the first time in an episode}. $N_e(s_t)$ is an episodic state counter that tracks exact state visitation counts, as defined by $(x, y)$ coordinates.

\paragraph{Measuring novelty with RND.} In smaller MDPs, it is possible to track exact state visitation counts, in which case the novelty is typically the inverse square root of visitation counts \cite{strehl2008analysis}. However, in larger environments where states are rarely revisited, we (like NovelD) use the popular Random Network Distillation (RND) \cite{burda2018exploration} technique as an approximate novelty measure. Specifically, the novelty of a state is measured by the prediction error of a state embedding network that is trained jointly with the agent to match the output of a fixed, random target network. The intuition is that states which the RND network has been trained on will have lower prediction error than novel states.

\subsection{Extension to L-NovelD}

Our incorporation of language is simple: we add an additional exploration bonus based on novelty defined according to the language descriptions of states:
\begin{align}
    \text{NovelD}_\ell(\ell_t, \ell_{t - 1}) \triangleq \max(N(\ell_{t}) - \alpha N(\ell_{t - 1}), 0) \cdot 
    \mathbbm{1}(N_e(\ell_t) = 1). \label{eq:lnoveld}
\end{align}
This bonus is identical to standard NovelD: $N(\ell)$ is the novelty of the description $\ell$ as measured by a \emph{separately parameterized} RND network encoding the description,\footnote{For multiple messages, we average NovelD of each messsage.\label{foot:noveld_multi_change}} and $N_e(\ell_t) = 1$ when the language description has been encountered for the first time this episode.
We keep the original NovelD exploration bonus, as language rewards may be sparse and a basic navigation bonus can encourage the agent to reach language-annotated states. The final intrinsic reward for L-NovelD is
\begin{equation}
    r^i_t = \text{L-NovelD}(s_t, s_{t - 1}, \ell_t, \ell_{t - 1}) \triangleq \text{NovelD}_s(s_t, s_{t - 1}) + \lambda_\ell \text{NovelD}_\ell(\ell_t, \ell_{t - 1})
\end{equation}
where $\lambda_\ell$ controls the trade-off between Equations~\ref{eq:noveld} and \ref{eq:lnoveld}.

One might ask why we do not simply include the language description $\ell$ as input into the RND network, along with the state. While this can work in some cases, decoupling the state and language novelties allow us to precisely control the trade-off between the two, with a hyperparameter that can be tuned to different tasks. In contrast, a combined input obfuscates the relative contributions of state and language to the overall novelty. Appendix~\ref{app:lnoveld_ablations} has ablations that show that (1) combining the state and language inputs or (2) using the language novelty term alone leads to worse performance.

\section{Experiments}
\label{sec:experiments}

We evaluate L-AMIGo, AMIGo, L-NovelD, and NovelD, implemented in the TorchBeast \cite{kuttler2019torchbeast} implementation of IMPALA \cite{espeholt2018impala}, a common asynchronous actor-critic method.
Besides vanilla IMPALA, we also compare to a naive (fixed) message reward given for any message in the environment, which is similar doing extrinsic reward shaping for all messages (e.g.\ LEARN \cite{goyal2019using}; also \cite{bahdanau2018learning,blukis2019learning,fu2019language,goyal2020pixl2r,schwartz2019language,waytowich2019narration}) or prior approaches that assume that messages are always helpful (\citet{harrison2018guiding}, ELLA \cite{mirchandani2021ella}); see Appendix~\ref{app:naive_reward} for more discussion on this baseline and its equivalencies to prior work.\footnote{For an implementation of a message reward with simple novelty-based decay, see the message-only L-NovelD ablation results in Appendix~\ref{app:lnoveld_ablations}, which underperforms full L-NovelD and L-AMIGo.}
We run each model 5 times across 13 tasks within two challenging procedurally-generated RL environments, MiniGrid \cite{chevalierboisvert2018minimalistic} and MiniHack \cite{samvelyan2021minihack}, and adapt baseline models provided for both environments \cite{campero2021learning,samvelyan2021minihack}; for full model, training, and hyperparameter details, see Appendix~\ref{app:full_details}.

\subsection{Environments}
\label{sec:environments}

\paragraph{MiniGrid.} Following \citet{campero2021learning}, we evaluate on the most challenging tasks in MiniGrid \cite{chevalierboisvert2018minimalistic}, which involve navigation and manipulation tasks in gridworlds: \textbf{KeyCorridorS\{3,4,5\}R3} (Figure~\ref{fig:language_overview}) and
\textbf{ObstructedMaze\_\{1Dl,2Dlhb,1Q\}}. These tasks involve picking up a ball in a locked room, with the key to the door hidden in boxes or other rooms and the door possibly obstructed. The suffix indicates the size of the environment, in increasing order.
See Appendix~\ref{app:full_minigrid_tasks} for more details.

To add language, we use the complementary BabyAI platform \cite{chevalierboisvert2019babyai}
which provides a grammar of 652 possible messages, involving \emph{goto}, \emph{open}, \emph{pickup}, and \emph{putnext} commands applied to a variety of objects qualified by type (e.g.\ \emph{box}, \emph{door}) and/or color (e.g.\ \emph{red}, \emph{blue}). The oracle language annotator emits a message when the corresponding action is completed. On average, only 6 to 12 messages (1-2\% of all 652) are needed to complete each task (see Appendix~\ref{app:full_minigrid_tasks} for all messages).

Note that since BabyAI messages are not included in the original environment from which we adapt baseline agents \cite{campero2021learning}, none of our MiniGrid agents encode language observations directly into the state. While it can be tempting and beneficial to use language in this way, one \emph{a priori} benefit of using language solely for exploration is that language is only needed during training, and not evaluation. Regardless, see Appendix~\ref{app:langstate} for additional experiments with MiniGrid agents that encode language into the state representation; while this boosts performance of baseline models, the experiments show that language-augmented exploration methods still outperform non-linguistic ones.

\paragraph{MiniHack.} MiniHack \cite{samvelyan2021minihack} is a suite of procedurally-generated tasks of varying difficulty set in the roguelike game NetHack \cite{kuttler2020nethack}. MiniHack contains a diverse action space beyond simple MiniGrid-esque navigation, including planning, inventory management, tool use, and combat. These actions cannot be expressed by $(x, y)$ positions, but instead are captured by in-game messages (Figure~\ref{fig:language_overview}). We evaluate our methods on a representative suite of tasks of varying difficulty: \textbf{River}, \textbf{Wand of Death (WoD)-\{Medium,Hard\}}, \textbf{Quest-\{Easy,Medium\}}, and \textbf{MultiRoom-\{N2,N4\}-Extreme}.

For space reasons, we describe the WoD-Hard environment here, but defer full descriptions of tasks (and messages) to Appendices~\ref{app:full_minihack_tasks} and \ref{app:full_minihack_messages}. In WoD-Hard, depicted in Figure~\ref{fig:language_overview}, the agent must learn to use a \emph{Wand of Death}, which can zap and kill enemies. This involves a complex sequence of actions: the agent must find the wand, pick it up, choose to \emph{zap} an item, select the wand in the inventory, and finally choose the zapping direction (towards the minotaur which is pursuing the player). It must then proceed past the minotaur to the goal to receive reward. Taking these actions out of order (e.g.\ trying to \emph{zap} with nothing in the inventory, or selecting something other than the wand) has no effect.

It is difficult to enumerate all MiniHack messages, as they are hidden in low-level game code which has many edge cases. As an estimate, we can examine expert policies: agents which have solved WoD tasks encounter around 60 messages, of which only 5--10 (8--16\%) are needed for successful trajectories, including inventory (\emph{f - a metal wand}, \emph{what do you want to zap?}, \emph{in what direction?}) and combat (\emph{You kill the minotaur!}, \emph{Welcome to level 2.}) messages; most are irrelevant (e.g.\ picking up and throwing stones) or nonsensical (\emph{There is nothing to pick up}, \emph{That is a silly thing to zap}). In the other tasks, only 8--18\% of the hundreds of unique messages are needed for success (Appendix~\ref{app:full_minihack_tasks}).

Unlike the MiniGrid environments, we adapt baseline models from \cite{samvelyan2021minihack}, which all already encode the in-game message into the state representation. Despite this, as we will show, using language as an explicit target for exploration outperforms using language as a state feature alone.

\section{Results}
\begin{figure}[t]
{
    \centering
    \includegraphics[width=\linewidth]{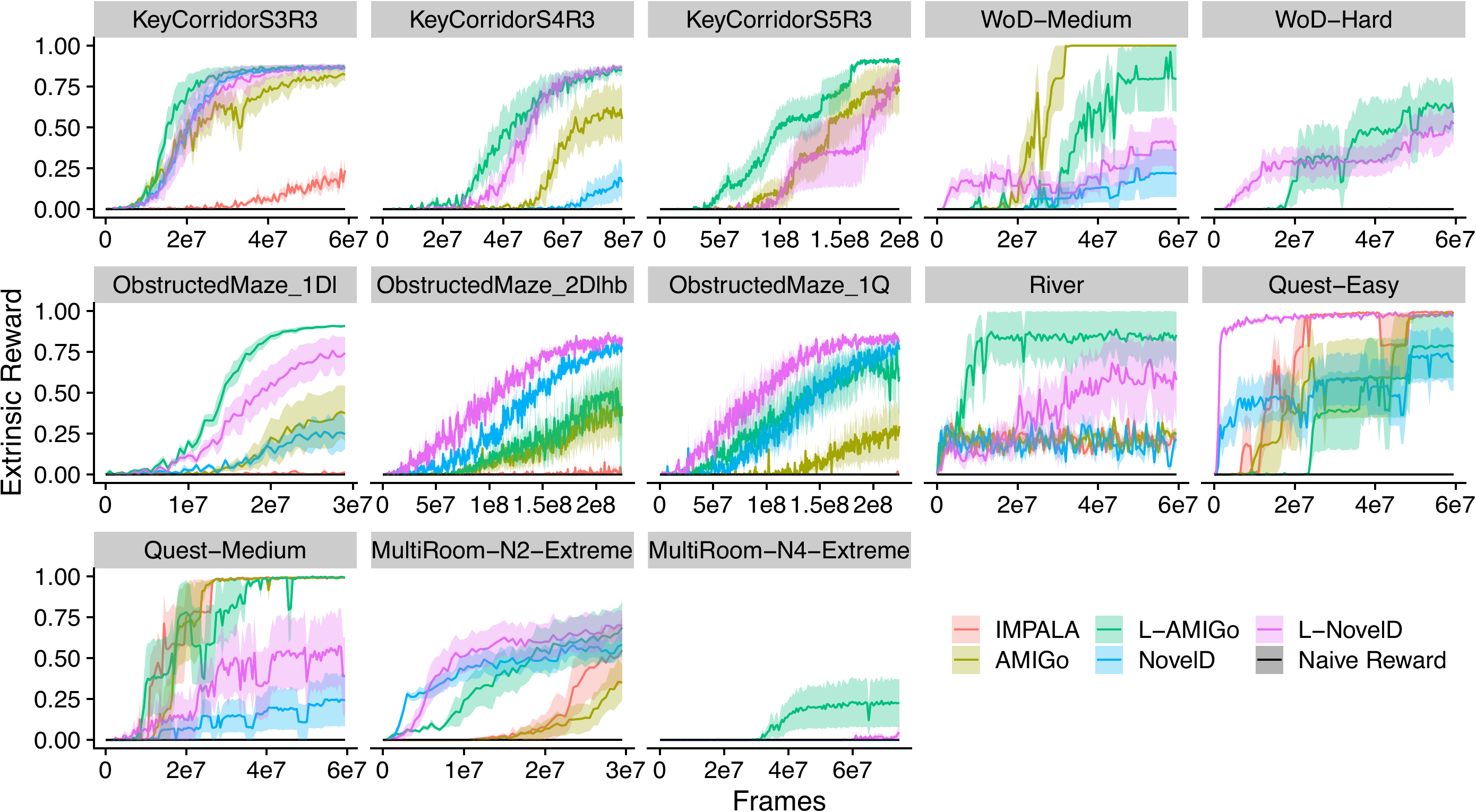}
    }
    \vspace{-1em}
    \caption{\textbf{Training curves}. Mean extrinsic reward ($\pm$ std err) across 5 independent runs for each model and environment. In general, linguistic variants outperform their non-linguistic forms.} 
    \label{fig:main_results}
    \vspace{-0.3em}
\end{figure}

\begin{figure}[t]
    \centering
     \begin{minipage}{0.48\textwidth}
         \centering
         \includegraphics[width=\linewidth]{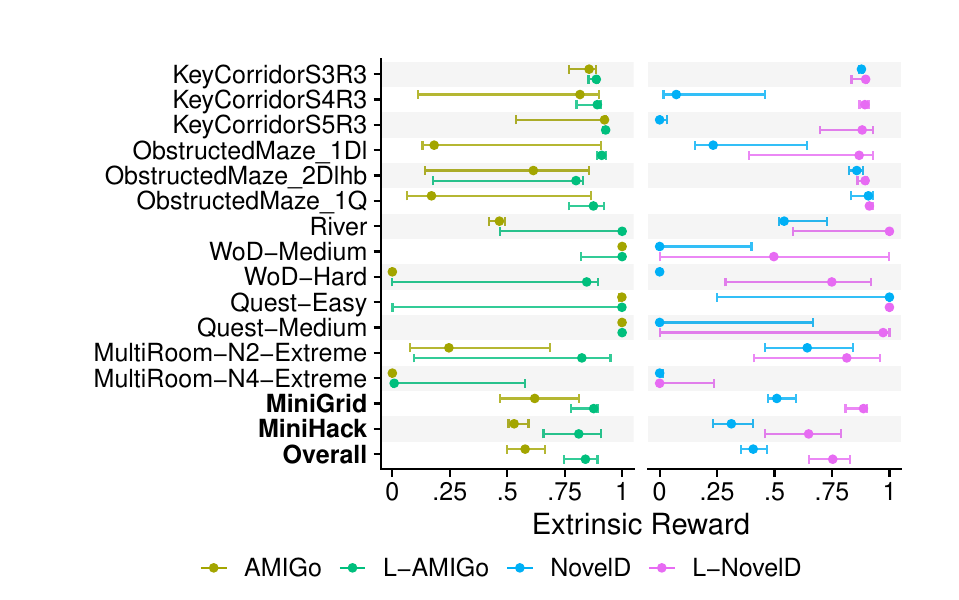}
         \caption{\textbf{Aggregate performance}. Interquartile mean (IQM) of models across tasks. Dot is median; error bars are 95\% bootstrapped CIs.}
         \label{fig:iqm}
     \end{minipage}\hfill
     \begin{minipage}{0.48\textwidth}
         \centering
         \includegraphics[width=\linewidth]{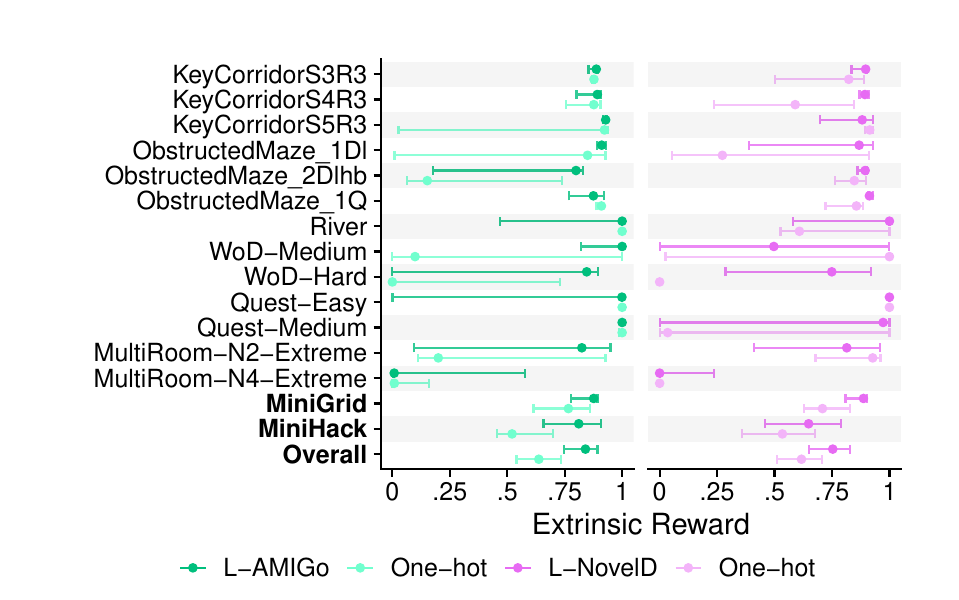}
         \caption{\textbf{One-hot performance.} Models compared to variants with one-hot non-compositional goals. Plot elements same as Figure~\ref{fig:iqm}}
         \label{fig:onehot}
     \end{minipage}
     \vspace{-1em}
\end{figure}
Figure~\ref{fig:main_results} shows training curves with AMIGo, NovelD, language variants, and the IMPALA and naive message reward baselines.
Following \citet{agarwal2021deep}, we summarize these results with the interquartile mean (IQM) of all methods in Figure~\ref{fig:iqm}, with bootstrapped 95\% confidence intervals constructed from 5k samples per model/env combination.\footnote{See Appendix~\ref{app:more_plots} for full numeric tables and area under the curve (AUC)/probability of improvement plots.}\footnote{See Appendix~\ref{app:ablations} for ablation studies of L-AMIGo's grounding network and the components of L-NovelD.}
We come to the following conclusions:

\textbf{Linguistic exploration outperforms non-linguistic exploration.} Both algorithms, L-AMIGo and L-NovelD, outperform their nonlinguistic counterparts. Despite variance in runs and across environments, averaged across all environments (\textbf{Overall}) we see a statistically significant improvement of L-AMIGo over AMIGo (.27 absolute, 47\% relative) and of L-NovelD over NovelD (.35 absolute, 85\% relative). In some tasks, Figure~\ref{fig:main_results} shows that L-AMIGo and L-NovelD reach the same asymptotic performance as their non-linguistic versions, but with better sample efficiency and stability (e.g.\ KeyCorridorS3R3 L-AMIGo, Quest-Easy L-NovelD; see Appendix~\ref{app:auc} AUC plots). Lastly, the failure of the naive message reward shows that indiscriminate reward shaping fails in tasks with sufficiently diverse language; instead, some notion of novelty or difficulty is needed to make progress.

\textbf{Linguistic exploration excels in larger environments.} Our tasks include sequences of environments with the same underlying dynamics, but larger state spaces and thus more challenging exploration problems. In general, larger environments result in bigger improvements of linguistic over non-linguistic exploration, since the space of messages remains relatively constant even as the state space grows.  For example, there is no difference in ultimate performance for language/non-language variants on KeyCorridorS3R3, yet the gaps grow as the environment size grows to KeyCorridorS5R3, especially in L-NovelD. A similar trend can be seen in the WoD tasks, where AMIGo actually outperforms L-AMIGo in WoD-Medium, but in WoD-Hard is unable to learn at all.

\subsection{Interpretability}

\begin{figure*}[t]
    \centering
     \begin{subfigure}[t]{0.41\linewidth}
         \includegraphics[width=\linewidth]{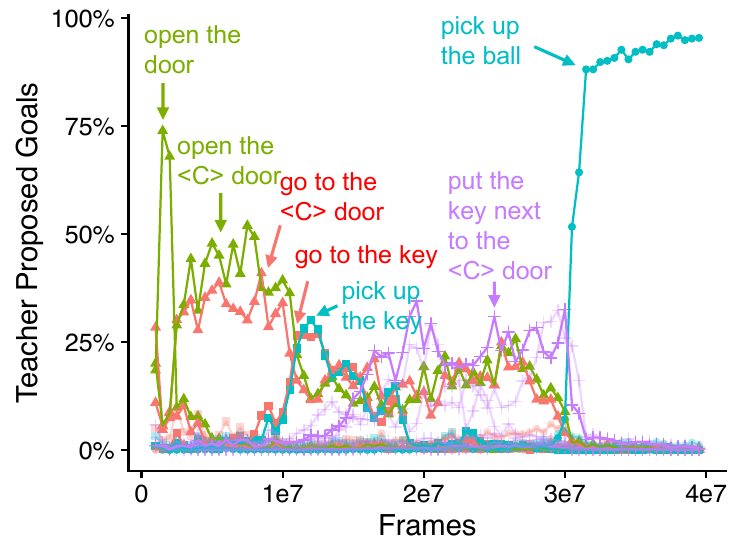}
         \caption{L-AMIGo, KeyCorridorS4R3}
     \end{subfigure}
     \begin{subfigure}[t]{0.5\linewidth}
         \includegraphics[width=\linewidth]{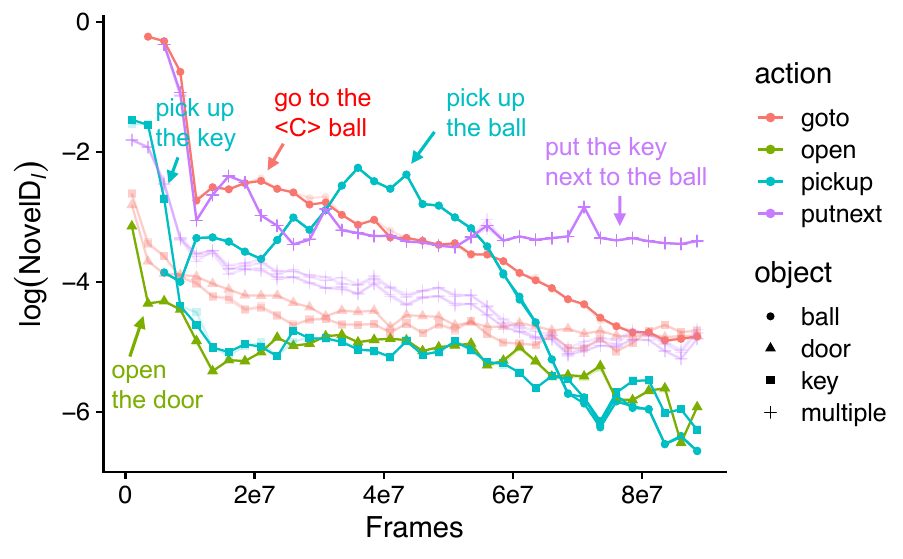}
         \caption{NovelD, KeyCorridorS4R3}
     \end{subfigure}
     \hfill
     \begin{subfigure}[b]{0.5\linewidth}
         \centering
         \vspace{0.3em}
         \includegraphics[width=\linewidth]{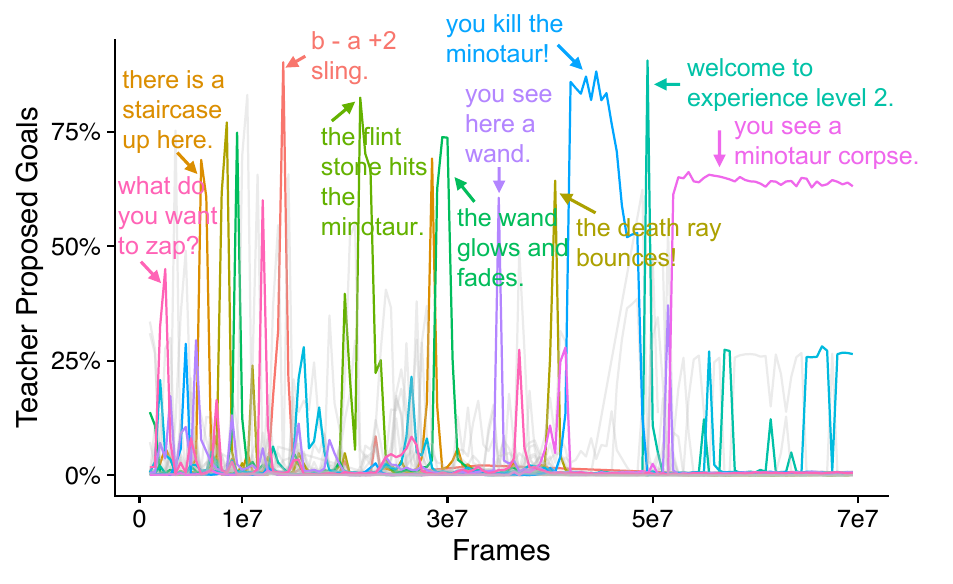}
         \caption{L-AMIGo, WoD-Hard}
     \end{subfigure}%
     ~
     \begin{subfigure}[b]{0.5\linewidth}
         \centering
         \vspace{0.3em}
         \includegraphics[width=\linewidth]{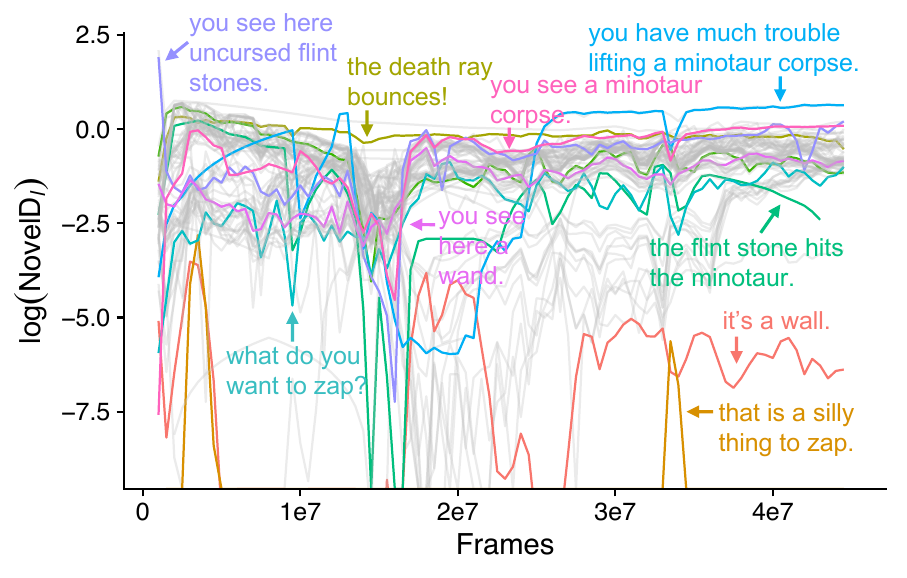}
         \caption{NovelD, WoD-Hard}
     \end{subfigure}
    \caption{\textbf{Interpretation of language-guided exploration.} For the KeyCorridorS4R3 and WoD-Hard environments, shown are curricula of goals proposed by the L-AMIGo teacher (a,c) and the intrinsic reward of messages (some examples labeled) for L-NovelD (b,d).}
    \label{fig:curricula}
    \vspace{-1.2em}
\end{figure*}
One auxiliary benefit of our language-based methods is that the language states and goals can provide insight into an agents' training and exploration process. We demonstrate how L-AMIGo and L-NovelD agents can be interpreted in Figure~\ref{fig:curricula}.

\paragraph{Emergent L-AMIGo Curricula.}
\citet{campero2021learning} showed that AMIGo teachers produce an interpretable curriculum, with initially easy $(x, y)$ goals located next to the student's start location, and later harder goals referencing distant locations behind doors. In L-AMIGo, we can see a similar curriculum emerge through the proportion of \emph{language goals} proposed by the teacher throughout training.
In the KeyCorridorS4R3 environment (Figure~\ref{fig:curricula}a), the teacher first proposes the generic goal \emph{open (any) door} before then proposing goals involving specific colored doors (where \texttt{<C>} is a color). Next, the agent discovers keys, and the teacher proposes \emph{pick[ing] up the key} and putting it in certain locations. Finally, the teacher and student converge on the extrinsic goal \emph{pick up the ball}.

Due to the complexity of the WoD-Hard environment, the curriculum for the teacher is more exploratory (Figure~\ref{fig:curricula}c). The teacher proposes useless goals at first, such as finding staircases and slings. At one point, the teacher proposes throwing stones at the minotaur (an ineffective strategy) before devoting more time towards wand actions (\emph{you see here a wand}, \emph{the wand glows and fades}). Eventually, as the student grows more competent, the teacher begins proposing goals that involve directly killing the minotaur (\emph{you kill the minotaur}, \emph{welcome to experience level 2}) before converging on the message \emph{you see a minotaur corpse}---the final message needed to complete the episode.

\paragraph{L-NovelD Message Novelty.}
Similarly, L-NovelD allows for interpretation by examining the messages with highest intrinsic reward as training progresses. In KeyCorridorS4R3 (Figure~\ref{fig:curricula}b), the novelty of easy goals such as \emph{open the door} decreases fastest, while the novelty of the true extrinsic goal (\emph{pick up the ball}) and even rarer actions (\emph{put the key next to the ball}) remains high throughout training. In WoD-Hard (Figure~\ref{fig:curricula}d), messages vary widely in novelty: simple and nonsensical messages like \emph{that is a silly thing to zap} and \emph{it's a wall} quickly plummet, while more novel messages are rarer states that require killing the minotaur (\emph{you have trouble lifting a minotaur corpse}).

\subsection{Do semantics matter?}
\label{sec:compositionality}
Language not only denotes meaningful features in the world; its lexical and compositional semantics also explain how actions and states relate to each other. For example, in L-AMIGo, an agent might more easily \emph{go to the red door} if it already knows how to \emph{go to the yellow door}. Similarly, in L-NovelD, training the RND network on the message \emph{go to the yellow door} could lower novelty of similar messages like \emph{go to the red door}, which might encourage exploration of semantically broader states.
While our primary focus is not on whether agents can generalize to new language instructions or states, we are still interested in whether these semantics improve exploration for extrinsic rewards.

To check this hypothesis, in Figure~\ref{fig:onehot} we run ``one-hot'' variants of L-AMIGo and L-NovelD where the semantics of the language annotations are hidden: each message is replaced with a one-hot identifier (e.g.\ \emph{go to the red door} $\rightarrow$ 1, \emph{go to the blue door} $\rightarrow$ 2) but otherwise functions identically to the original message. We make two observations. \textbf{(1)} One-hot goals actually perform quite competitively, demonstrating that the primary benefit of language in these tasks is to abstract over the state space, rather than provide fine-grained semantic relations between states. \textbf{(2)} Nevertheless, L-AMIGo is able to exploit semantics, with a significant improvement (.20 absolute, 32\% relative) in aggregate performance over one-hot goals, in contrast to L-NovelD, which shows no significant difference.
We leave for future work a more in-depth investigation into what kinds of environments and models might benefit more from language semantics.
\section{Discussion}
\label{sec:discussion}
The key insight in this paper is that language, even if noisy and often unrelated to the goal, is a more abstract, efficient, and interpretable space for exploration than state representations. To support this, we have presented variants of two popular state-of-the-art exploration methods, L-AMIGo and L-NovelD, that outperform their non-linguistic counterparts by 47--85\% across 13 language-annotated tasks in the challenging MiniGrid and MiniHack environment suites.

Despite their success here, our models have some limitations. First, as is common in work like ours, it will be important to alleviate the restriction on oracle language annotations, perhaps by using learned state description models \cite{mirchandani2021ella,tam2022semantic}. Furthermore, L-AMIGo specifically cannot handle tasks such as the full NetHack game which have unbounded language spaces and many redundant goals (e.g.\ \emph{go to/approach/arrive at the door}), since it selects a single goal which must be achieved verbatim. An exciting extension to L-AMIGo would propose abstract goals (e.g.\ \emph{kill [any] monster} or \emph{find a new item}), possibly in a continuous semantic space, that can be satisfied by multiple messages.

More general extensions include better understanding when and why language semantics benefits exploration (Section~\ref{sec:compositionality}) and using pretrained models to imbue semantics into the models beforehand \cite{tam2022semantic}. Additionally, although the agents in this work are able to explore even when not all language is useful, we must take caution in adversarial settings where the language is completely unrelated to the extrinsic task (and thus useless) or even describes harmful behaviors. Future work should measure how robust these methods are to the noisiness and quality of the language.
Nevertheless, the success of L-AMIGo and L-NovelD demonstrates the power of even noisy language in these domains, underscoring the importance of abstract and semantically-meaningful measures of exploration in RL.

\begin{ack}
We thank Heinrich K\"{u}ttler, Mikael Henaff, Andy Shih, Alex Tamkin, and anonymous reviewers for constructive comments and feedback, and Mikayel Samvelyan for help with MiniHack. JM is supported by an Open Philanthropy AI Fellowship.
\end{ack}

{
\small
\bibliographystyle{abbrvnat}
\bibliography{lamigo}
}

\newpage
\appendix

\renewcommand\thefigure{S\arabic{figure}}
\renewcommand\thetable{S\arabic{table}}
\renewcommand\thealgorithm{S\arabic{algorithm}}
\setcounter{figure}{0}
\setcounter{table}{0}
\setcounter{algorithm}{0}

\section{L-AMIGo algorithm}
\label{app:lamigo_algorithm}

\begin{algorithm}[ht]
    \caption{Asynchronous learning step for L-AMIGo}
    \label{alg:lamigo}
    \begin{algorithmic}[1]
        \State \textbf{Input:} student batch size $B_S$, teacher policy batch size $B_\text{policy}$, grounding network batch size $B_{\text{ground}}$
        \State $\mathcal{B}_\text{policy} \gets \varnothing, \mathcal{B}_\text{ground} \gets \varnothing$ \Comment{Init teacher batches}
        \State $\mathcal{G} \gets \varnothing$ \Comment{Track descriptions seen thus far}
        \While{not converged}
            \State Sample batch $\mathcal{B}$ of size $B_S$ from actors
            \State Train student on $\mathcal{B}$
            \State Add new descriptions $\ell_t$ in the batch to $\mathcal{G}$
            \State Update $\mathcal{B}_{\text{policy}}$ with $(s_0, \ell_t, r^T_t)$ tuples where goal $\ell_t$ completed ($r^T_t = +\alpha)$ or episode ended ($r^T_t = -\beta$)
            \If{$| \mathcal{B}_\text{policy} | > B_\text{policy}$}
                \State Train teacher policy on $\mathcal{B}_\text{policy}$; $\mathcal{B}_\text{policy} \gets \varnothing$
            \EndIf
            \State Update $\mathcal{B}_\text{ground}$ with ($s_0, \ell_{\text{1st}}$) tuples 
            \If{$| \mathcal{B}_\text{ground} | > B_\text{ground}$}
                \State Train grounding net on $\mathcal{B}_\text{ground}$; $\mathcal{B}_\text{ground} \gets \varnothing$
            \EndIf
        \EndWhile
    \end{algorithmic}
\end{algorithm}

Algorithm~\ref{alg:lamigo} describes the joint student/teacher learning step of L-AMIGo. Batches of experience $\mathcal{B}$ are generated from actors, which are used to update the student policy, and the teacher policy at different intervals defined by batch sizes $B_{\text{policy}}$ and $B_{\text{ground}}$. The set of goals $\mathcal{G}$ known to the teacher is also progressively updated.

To reiterate the teacher training process: the teacher policy network is trained on tuples of student initial state $s_0$, proposed goal $\ell_t$, and teacher rewards $r^T_t$ (which is $+\alpha$ if the goal was completed by the student in $\geq t^*$ steps, or $-\beta$ if the goal was completed in $< t^*$ steps or never completed before episode termination). The teacher grounding network is trained on tuples of initial state $s_0$ and the first language description encountered along that trajectory $\ell_\text{1st}$. Given $s_0$, the grounding network is asked to predict 1 for $\ell_\text{1st}$ and 0 for all other goals known to the teacher at the time.

\section{Details on naive message reward and equivalencies to prior work}
\label{app:naive_reward}

As discussed in Section~\ref{sec:related-work}, prior approaches to language-guided reward shaping assume either a single extrinsic goal, or that language annotations are always helpful for progress in the environment. Here we show how these methods can all be approximately captured by a naive message reward, which our experiments show is insufficient for learning in sufficiently large linguistic spaces (Figure~\ref{fig:main_results}).

\begin{itemize}
    \item \citet{harrison2018guiding} (also \cite{tasrin2021influencing}) propose a policy shaping method that generates language annotations for (action, state) pairs from simulated oracles, then for each language annotation learns a ``critique policy'' that mimics the human action distributions for the current state and language annotation. This is used to update the prior action distribution of the learned policy during training: given a state, the most likely language annotation over all possible actions is inferred (via Bayes' rule); then the critique policy for this language annotation is mixed in with the current policy's action distribution via a tunable hyperparameter. While the focus is on policy shaping rather than reward shaping, agents are simply pushed in the direction of the most salient language annotation, and there are no preferences for or against certain language. The overall effect, then, is that policies are pushed indiscriminately towards encountering \emph{any} message in the environment, which is similar to a fixed message reward.
    
    A further limitation of \citet{harrison2018guiding} that prevents straightforward application here is that it requires sufficient training data for training a critique policy offline \emph{for every possible message encountered in the environment}. In contrast, we assume no such pretraining data; in fact, we have no knowledge about what messages we might encounter at all.
    \item ELLA \cite{mirchandani2021ella} is closer to our setting, since the authors acknolwedge that not all messages or subgoals in a trajectory may be relevant for an overall goal, especially in the multi-task settings they explore where the extrinsic goal is subject to change. Accordingly, ELLA proposes to learn a ``Relevance Classifier'' that predicts whether or not a message is useful for the extrinsic goal by training on the messages encountered along trajectories that resulted in positive reward. However, in our setting, we have no extrinsic goals, and in most of our tasks, random exploration fails to attain \emph{any} positive trajectories to provide such training data for a classifier (Figure~\ref{fig:main_results}, IMPALA curves). We could implement ELLA with an uninformative reference classifier, i.e.\ one that assumes all messages are relevant for the extrinsic goal. Then ELLA reduces to simply giving a fixed reward for any message encountered.
    \item Finally, a large class of reward shaping and inverse RL methods operate primarily by giving rewards associated with a (linguistic) extrinsic goal \cite{bahdanau2018learning,blukis2019learning,fu2019language,goyal2019using,goyal2020pixl2r,harrison2018guiding,mirchandani2021ella,schwartz2019language,tasrin2021influencing,waytowich2019narration}. As one representative example,
    LEARN \cite{goyal2019using} proposes to train a model to associate an action at a given timestep with the probability it is \emph{related} to an extrinsic language command: $p_R(a_t, \ell)$, as well as the complementary probability that an action is \emph{unrelated}: $p_U(a_t, \ell) = 1 - p_R(a_t, \ell)$. The difference in these probabilities can thus be used as a reward signal: $r^i_t = p_R(a_t, \ell) - p_U(a_t, \ell)$.
    
    However, in our case we have no extrinsic instruction; rather, we have many intermediate low-level messages of unknown relevance to the extrinsic goal. A naive solution for LEARN-style approaches is thus to do reward shaping for every intermediate message, i.e.\ assign rewards whenever any annotation $\ell$ is deemed relevant. Moreover, we do not need to predict when an action is relevant to $\ell$; the frequent annotations we receive already confirm that an action is linguistically relevant. Thus, in our setting, we can set $p_R(a, \ell) = 1$ if $\ell$ is observed in the given state, and $p_R(a, \ell) = 0$ otherwise.\footnote{An even stricter interpretation would be to give partial rewards even for states with the null message $\varnothing$ by training a classifier to predict relevance for any non-null linguistic state. However, intermediate message rewards are already fairly common in our setting (it is easy to walk to the nearest wall in MiniHack, for example), and this would not solve the fundamental problem that without learning to disprefer certain messages, an agent will be stuck pursuing locally optimal messages.} Using this as the reward is equivalent to the baseline employed here. Technically, LEARN proposes a slightly modified reward $r'^{i}_t = \gamma r^i_t - r^i_{t - 1}$, i.e.\ the difference in reward between timesteps where $\gamma$ is the MDP discount factor, but this made no difference in our experiments; thus, for brevity we report the single fixed reward.
\end{itemize}

For the naive reward baseline reported in Figure~\ref{fig:main_results}, we performed a grid search on the intrinsic reward coefficient $\lambda \in \{0.1, 0.5, 1.0\}$, though modifying $\lambda$ made no difference and results are reported with $\lambda = 0.1$. In all cases, agents trained with such rewards fail because without some notion of message novelty or difficulty, the agents are stuck exploiting easy-to-achieve, locally-optimal messages (e.g. running into the nearest wall in MiniHack, or the nearest door in MiniGrid); in fact, rewards for easy messages \emph{discourage} exploration and lead to worse reward than even the vanilla IMPALA baselines! Thus, some way of measuring novelty/progress in the space of language annotations must be used, which is operationalized in L-AMIGo (via the continually growing difficulty threshold given to the teacher) and L-NovelD (via the decaying reward given for seeing the same message over and over again), though note that for L-NovelD, using a message reward with novelty-based decay is insufficient (Appendix~\ref{app:lnoveld_ablations}).

\section{Architecture and training details}
\label{app:full_details}

Here, we describe the architecture, training details, and hyperparameters for the MiniGrid and MiniHack tasks. Our code is available in the supplementary material and also at \url{https://anonymized}.

\subsection{MiniGrid}

\begin{figure}[ht]
    \centering
    \includegraphics[width=\linewidth]{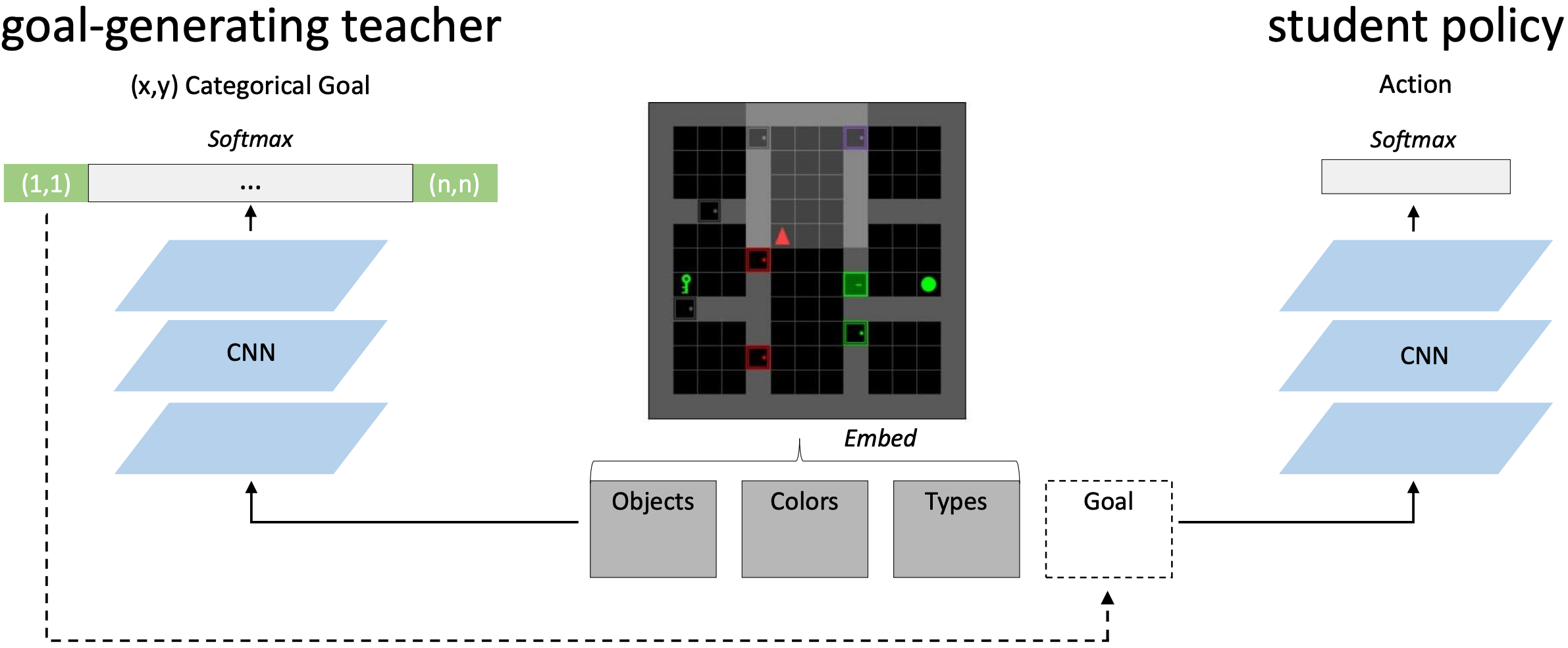}
    \caption{\textbf{Original AMIGo model overview.} Original figure from \citet{campero2021learning}, reproduced with permission.}
    \label{fig:amigo}
    \vspace{-1em}
\end{figure}

All models are adapted from \citet{campero2021learning}. Figure~\ref{fig:amigo} from \citet{campero2021learning} details the architecture of the standard AMIGo student and teacher.

\paragraph{AMIGo student.} The student gets an $7 \times 7 \times 3$ partial and egocentric grid representation, where each cell in the $7 \times 7$ grid is represented by 3 features indicating the type of the object, color of object, and object state (e.g.\ for doors, open/closed). The student first embeds the features into type/color/state embeddings of size 5, 3, and 2 respectively, as well as the $(x, y)$ goal as a singleton feature, to get a $7 \times 7 \times 11$ grid which is then fed through a 5-layer convolutional neural network interleaved with Exponential Linear Units (ELUs; \citealp{clevert2016fast}). Each layer has 32 output channels, filter size of $(3, 3)$, 2 stride, and 1 padding. The output of the ConvNet is then flattened and fed through a fully-connected layer to produce a 256-d state embedding. The policy and value functions are linear layers on top of this state embedding. The policy in particular produces a distribution over 7 possible actions (forward, pick up, put down, open, left, right, and a ``done'' action).

\paragraph{AMIGo teacher.} The teacher gets an $N \times N \times 3$ fully observed grid encoding of the environment, where $N$ varies according to the environment size. Like the student, the teacher embeds the grid representation into embeddings which are then fed through a 4-layer dimensionality-preserving neural network again interwoven with ELUs. Each convolutional layer has 16 output channels, filter size of $(3, 3)$, 1 stride, and 1 padding, except the last layer which has only 1 output channel. The ConvNet processes the input embeddings into a spatial action distribution over grid locations, from which an goal is sampled. The value head takes as input the penultimate layer from the ConvNet (i.e.\ a grid of $N \times N \times 16$) to produce the value estimate.

\paragraph{L-AMIGo Student.} The L-AMIGo student has the same ConvNet base as the standard AMIGo student, but without an $(x, y)$ goal channel feature (so the input to the ConvNet is $7 \times 7 \times 10$ instead of 11). To encode the goal, the student uses a one-layer Gated Recurrent Unit (GRU; \citet{cho2014learning}) recurrent neural network (RNN) with embedding size 64 and hidden size 256. The last hidden state of the GRU is taken as the goal representation, which is then concatenated with the state embedding to be fed into the policy and value functions, respectively.

\paragraph{L-AMIGo Teacher.} The L-AMIGo \emph{policy network} has the same ConvNet as the standard AMIGo teacher, but without the last layer (so the output is a $N \times N \times 16$ grid). The last output of the ConvNet is averaged across the spatial map to form a final 16-dimensional state embedding. The L-AMIGo teacher also has a GRU of same dimensionality as the L-AMIGo student. To propose a goal, the teacher embeds each known goal in the vocabulary with the GRU to produce 256-dimensional goal embeddings which are then projected via a linear layer into the 16-dimensional goal embedding space. The dot product of the goal embeddings and the state form the logits of the goal distribution.
The value head takes as input the $N \times N \times 16$ unaveraged state representation concatenated with the logits of the distribution of language goals.

The \emph{grounding network} also uses the 16-dimensional state embedding and the same GRU as the L-AMIGo policy network, but the 256-d goal embeddings are projected via a separate linear layer to a separate set of 16-dimensional embeddings. The dot product between these grounder-specific goal embeddings and the state embedding represent the log probabilities of the goal being achievable in an environment.

\paragraph{NovelD and L-NovelD.}

For NovelD experiments we use the student policy of L-AMIGo but with the goal embedding always set to the 0 vector.

The NovelD RND network uses the same convolutional network as the AMIGo/L-AMIGo students, taking in an egocentric agent view. The L-NovelD message embedding network uses a GRU parameterized identically to those of the L-AMIGo student and teacher. For NovelD experiments we set $\alpha = 1$, scale the RND loss by 0.1, and use the same learning rate as the main experiments. Using a grid search, we found optimal scaling factors of the standard NovelD reward to be 0.5 and the L-NovelD reward (i.e.\ $\lambda_\ell$) also to be 0.5.

\paragraph{Hyperparameters.}
We use the same hyperparameters as in the original AMIGo paper: a starting difficulty threshold $t^*$ of 7, a maximum difficulty $t^*$ of 100, positive reward for the teacher $+ \alpha = 0.7$, negative reward $-\beta = -0.3$, learning rate $10^{-4}$ which is linearly annealed to 0 throughout training, batch size $32$, teacher policy batch size $32$, teacher grounder batch size $100$, unroll length $100$, RMSprop optimizer with $\varepsilon = 0.01$ and momentum $0$, entropy cost $0.0005$, generator entropy cost $0.05$, value loss cost $0.5$, intrinsic reward coefficient $\lambda = 1$.

\subsection{MiniHack}

Models are adapted from baselines established for the NetHack \cite{kuttler2020nethack} and MiniHack \cite{samvelyan2021minihack} Learning Environments.

\paragraph{AMIGo student.} The student is a recurrent LSTM-based \cite{hochreiter1997long} policy. The NetHack observation contains a $21 \times 79$ matrix of glyph identifiers, a 21-dimensional feature vector of agent stats (e.g.\ position, health, etc), and a 256-character (optional) message. The student produces 4 dense representations from this observation which are then concatenated to form the state representation.

\textbf{First}, an embedding of the entire game area is created. Each glyph is converted into a $64$-dimensional vector embedding, i.e. the input is now a $21 \times 79 \times 64$ grid. This entire grid is fed through a 5-layer ConvNet interleaved with ELUs. Each conv layer has a filter size of $(3, 3)$, stride 1, padding 1, and 16 output channels, except for the last which has 8 output channels. \textbf{Second}, an embedding of a $9 \times 9$ egocentric crop of the grid around the agent is created. This is created by feeding the crop through another separately-parameterized ConvNet of the same architecture. \textbf{Third}, the 21-d feature vector of agent stats is fed into a 2-layer MLP (2 layers with 64-d outputs with ReLUs in between) to produce a 64-d representation of the agent stats. \textbf{Fourth}, the message is parsed with the BERT \cite{devlin2019bert} WordPiece tokenizer and fed into a GRU with embedding size 64 and hidden size 256. These four embeddings are then concatenated and form the state representation, which is then fed into the LSTM which contextualizes the current representation. This final representation is fed into linear policy and value heads to produce action distribution and value estimates.

\paragraph{AMIGo teacher.} The teacher first embeds the agent stats using the same MLP that the student uses. It has the same 5-layer ConvNet used by the student to embed the full game area, except the input channels are modified to accept the 64-d feature vector concatenated to every cell in the $21 \times 79$ grid, so the input to the ConvNet is $21 \times 79 \times 128$. Additionally, the output layer has only 1 output channel. This produces a spatial action distribution over the $21 \times 79$ grid from which a ``goal'' as $(x, y)$ coordinate is sampled.

\paragraph{L-AMIGo student.} The student works identically to the standard AMIGo student, except without the teacher goal channel concatenated into the grid input. Also, in addition to the 4 representation constructed from the observation, the student gets 2 more representations constructed from the teacher's language goal. \textbf{First}, the student uses the same GRU used to process the game message to encode the teacher's language goal to create a 256-d representation. \textbf{Second}, the difference between the observed language embedding and the teacher language embedding is used as an additional feature (when this is the 0 vector then the student has reached the goal). All together, this forms 6 representations that together constitute the state representation.

\paragraph{L-AMIGo teacher.} The teacher constructs the same state representation as the standard AMIGo student (\emph{not} the AMIGo teacher). The teacher then embeds all known goals with a word-level GRU with the same architecture as the message network used by the student. These 256-d message embeddings are then projected to the state representation hidden size, and the dot product between message embeddings and state representations forms the distribution over goals. Similarly, for the grounding network, the 256-d message embeddings are projected via a separate linear layer to produce probabilities of achievability. Like in MiniGrid, the state representation and the goal logits are used as input for the value head. 

\paragraph{NovelD and L-NovelD.} As in MiniGrid experiments, for NovelD experiments we use the L-AMIGo student policy where the goal embedding is always 0.

The NovelD RND network uses the cropped ConvNet representation of the student, fed through a linear layer to produce a 256-d state representation. Egocentric crops change more over time and are thus a more reliable signal of ``novelty'' than the full grid representation (verified with experiments).
Additionally, as done in \citet{burda2018exploration}, two more additional layers are added to the final MLP of the predictor network only (not the random target network). The L-NovelD message RND network uses the same word-level GRU architecture of the student.
We set $\alpha = 0.5$, scale the RND loss by 0.1, and use the same learning rate as the main student policy. The standard NovelD reward is left alone, and the L-NovelD reward hyperparameter ($\lambda_\ell$) is grid-searched and set to 50 for all MiniHack experiments except Quest-\{Easy,Medium\}, where it is 30.

\paragraph{Hyperparameters.}
We generally stick to the same hyperparameters of \citet{samvelyan2021minihack}. We use a starting difficulty threshold $t^*$ of 1, a maximum difficulty $t^*$ of 2 (a high goal difficulty is not as important for MiniHack), positive reward for the teacher $+ \alpha = 0.7$, negative reward $-\beta = -0.3$, linearly-annealed learning rate $10^{-4}$, batch size $32$, teacher policy batch size $32$, teacher grounder batch size $500$, unroll length $100$, RMSprop optimizer with $\varepsilon = 0.01$ and momentum $0$, entropy cost $0.0005$, generator entropy cost $0.05$, value loss cost $0.5$, intrinsic reward coefficient $\lambda = 0.4$. $\varepsilon$-greedy exploration was used for the teacher policy with $\varepsilon = 0.05$, which we found helped learning.

\subsection{Compute Details}
\label{app:compute_details}

Each model was run for 5 independent seeds on a machine in an independent cluster with 40 CPUs, 1 Tesla V100 GPU, and 64GB RAM. Runs take between 4 hours (for ObstructedMaze\_1Dl) to 20 hours (for the longest MultiRoom-N4-Extreme and KeyCorridorS5R3 tasks).

\section{Additional tables visualizations of main results}
\label{app:more_plots}

\subsection{Full numeric tables}

Table~\ref{tab:full_numeric_table} contains full numbers for IQM performance. This is the same data as Figure~\ref{fig:iqm}, just summarized in numeric form.

\begin{table}[ht]
    \small
    \centering
    \caption{\textbf{Full IQM numbers.} IQM performance ($\pm$ 95\% bootstrapped CIs) for models across tasks.}
    \vspace{1em}
    \begin{tabular}{lrrrr}
    \toprule
    Environment & AMIGo & L-AMIGo & NovelD & L-NovelD \\
    \midrule
KeyCorridorS3R3 & 0.86 (0.77, 0.89) & 0.89 (0.85, 0.90) & 0.88 (0.87, 0.88) & \textbf{0.90} (0.83, 0.90) \\ 
  KeyCorridorS4R3 & 0.82 (0.11, 0.90) & \textbf{0.89} (0.80, 0.91) & 0.07 (0.02, 0.45) & \textbf{0.89} (0.87, 0.91) \\ 
  KeyCorridorS5R3 & 0.92 (0.54, 0.93) & \textbf{0.93} (0.92, 0.93) & 0.00 (0.00, 0.03) & 0.88 (0.70, 0.93) \\ 
  ObstructedMaze\_1Dl & 0.18 (0.13, 0.91) & \textbf{0.91} (0.89, 0.93) & 0.23 (0.15, 0.64) & 0.87 (0.39, 0.93) \\ 
  ObstructedMaze\_2Dlhb & 0.61 (0.14, 0.86) & 0.80 (0.18, 0.83) & 0.86 (0.82, 0.88) & \textbf{0.89} (0.86, 0.90) \\ 
  ObstructedMaze\_1Q & 0.17 (0.06, 0.86) & 0.88 (0.77, 0.92) & \textbf{0.91} (0.83, 0.93) & \textbf{0.91} (0.91, 0.93) \\ 
  River & 0.47 (0.42, 0.49) & \textbf{1.00} (0.47, 1.00) & 0.54 (0.52, 0.73) & \textbf{1.00} (0.58, 1.00) \\ 
  WoD-Medium & \textbf{1.00} (1.00, 1.00) & \textbf{1.00} (0.82, 1.00) & 0.00 (0.00, 0.40) & 0.50 (0.00, 1.00) \\ 
  WoD-Hard & 0.00 (0.00, 0.00) & \textbf{0.85} (0.00, 0.90) & 0.00 (0.00, 0.00) & 0.75 (0.29, 0.92) \\ 
  Quest-Easy & \textbf{1.00} (0.99, 1.00) & \textbf{1.00} (0.00, 1.00) & \textbf{1.00} (0.25, 1.00) & \textbf{1.00} (1.00, 1.00) \\ 
  Quest-Medium & \textbf{1.00} (0.99, 1.00) & \textbf{1.00} (1.00, 1.00) & 0.00 (0.00, 0.67) & 0.97 (0.00, 1.00) \\ 
  MultiRoom-N2-Extreme & 0.25 (0.08, 0.69) & \textbf{0.82} (0.09, 0.95) & 0.64 (0.46, 0.84) & 0.81 (0.41, 0.96) \\ 
  MultiRoom-N4-Extreme & 0.00 (0.00, 0.00) & \textbf{0.01} (0.01, 0.58) & 0.00 (0.00, 0.01) & 0.00 (0.00, 0.24) \\ 
  \midrule
  \textbf{MiniGrid} & 0.62 (0.47, 0.81) & 0.88 (0.78, 0.89) & 0.51 (0.47, 0.59) & \textbf{0.89} (0.81, 0.90) \\ 
  \textbf{MiniHack} & 0.53 (0.51, 0.59) & \textbf{0.81} (0.65, 0.91) & 0.31 (0.23, 0.41) & 0.65 (0.46, 0.79) \\ 
  \textbf{Overall} & 0.57 (0.50, 0.66) & \textbf{0.84} (0.74, 0.89) & 0.41 (0.35, 0.47) & 0.76 (0.66, 0.83) \\ 
    \bottomrule
    \end{tabular}
    \label{tab:full_numeric_table}
\end{table}

\subsection{Probability of improvement} 
\label{app:p_improvement}

\begin{figure}[t]
    \centering
    \begin{minipage}[t]{0.48\textwidth}
        \centering
        \includegraphics[width=\linewidth]{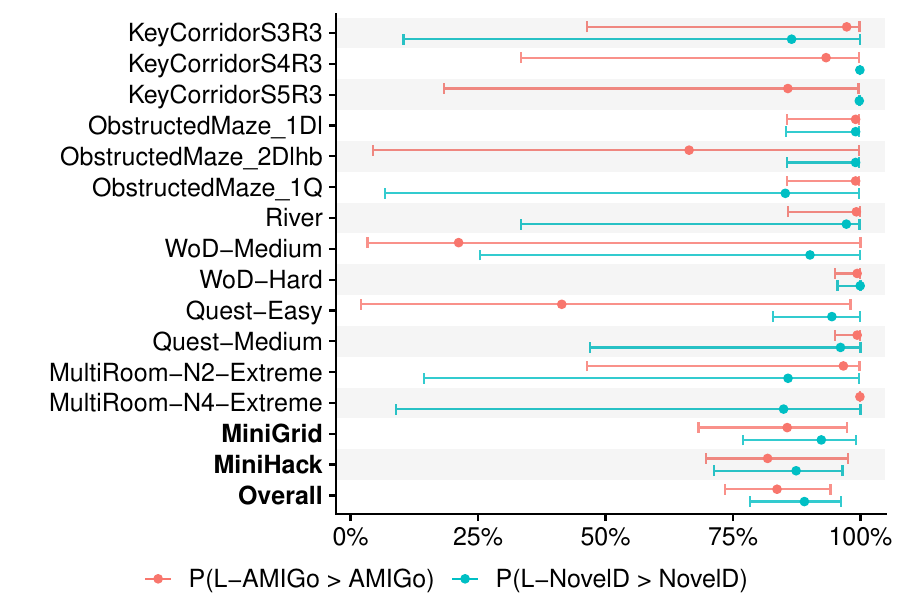}
        \caption{\textbf{Probability of improvement.} Probability of improvement of L-AMIGo over AMIGo, and L-NovelD over NovelD, as measured by Mann-Whitney U tests between their final performances. Plot elements same as Figure~\ref{fig:iqm}.}
        \label{fig:p_improvement}
    \end{minipage}
    \hfill
    \begin{minipage}[t]{0.48\textwidth}
        \centering
        \includegraphics[width=\linewidth]{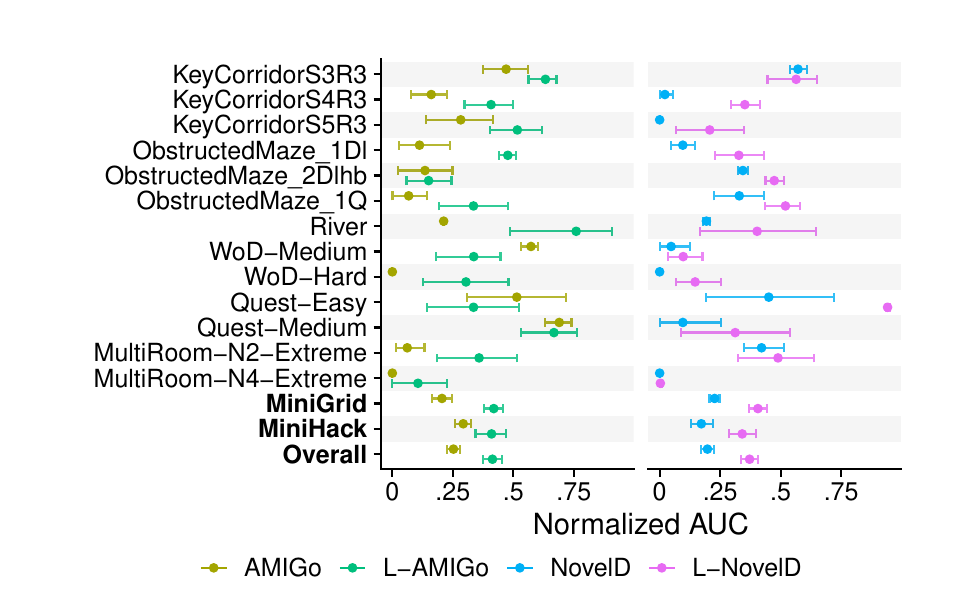}
        \caption{\textbf{Normalized AUC.} Normalized area under the curve (AUC) for AMIGo, L-AMIGo, NovelD, and L-NovelD. Plot elements same as Figure~\ref{fig:iqm}.}
        \label{fig:auc}
        \vspace{-1em}
    \end{minipage}
    \vspace{-1em}
\end{figure}

In addition to the IQM, another alternative interpretation of results as advocated in \citet{agarwal2021deep} is to use the \emph{probability of improvement} of algorithm $A$ over algorithm $B$, as measured by the nonparametric Mann-Whitney U test between independent runs from both algorithms. Figure~\ref{fig:p_improvement} shows results from such an evaluation, again evaluated over bootstrapped confidence intervals constructed from 5000 samples per model/env combination. Qualitative results are the same as in Figure~\ref{fig:iqm}: overall, across environments, L-AMIGo and L-NovelD are both highly likely to outperform AMIGo and NovelD.

\subsection{Area Under the Curve (AUC).}
\label{app:auc}
The IQM (Figure~\ref{fig:iqm}) and probability plots (Figure~\ref{fig:p_improvement}) use point estimates of ultimate performance attained after a fixed compute budget, which does not measure differences in sample efficiency between runs.  
As a measure which also elucidates differences in sample efficiency, we to use the normalized Area Under the Curve (AUC) to compare runs, as used in prior work \cite{goyal2019using}. Results are in Figure~\ref{fig:auc}, and are qualitatively similar to the IQM and probability plots, but here, differences in sample efficiency can be seen in some environments, e.g.\ in KeyCorridorS3R3 for L-AMIGo, which are absent from the IQM plot.

\section{MiniGrid experiments with language encoded into the state representation}
\label{app:langstate}

\begin{figure}[t]
    \begin{minipage}[t]{0.48\textwidth}
        \centering
        \includegraphics[width=\linewidth]{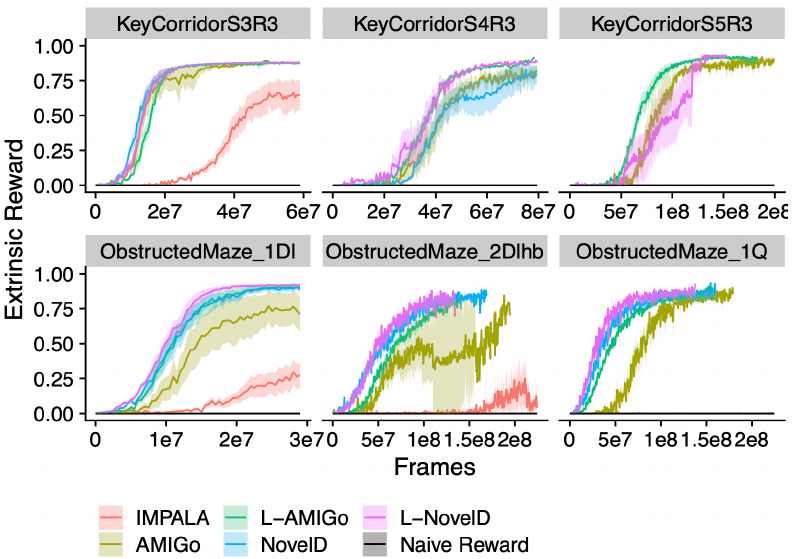}
        \caption{\textbf{Training curves for MiniGrid with language states}. Plot elements same as Figure~\ref{fig:main_results}.}
        \label{fig:main_results_langstate}
    \end{minipage}
    \hfill
    \begin{minipage}[t]{0.48\textwidth}
        \centering
        \includegraphics[width=\linewidth]{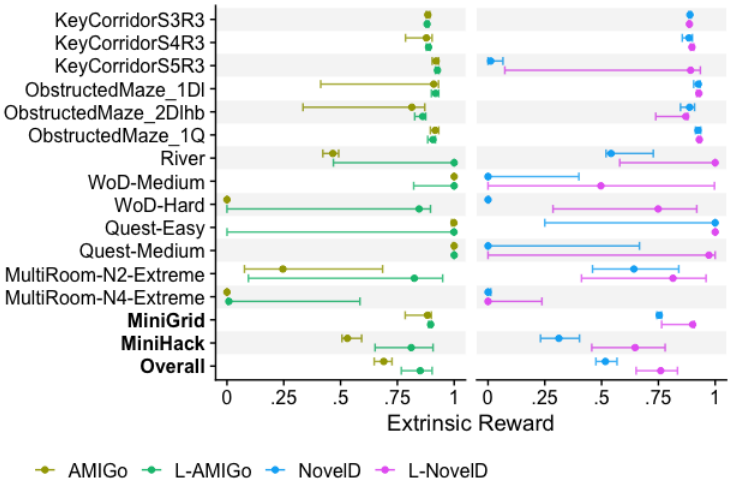}
        \caption{\textbf{IQMs with language states}. Plot elements same as Figure~\ref{fig:iqm}.}
        \label{fig:iqm_langstate}
    \end{minipage}
    \vspace{-1em}
\end{figure}

Here we run MiniGrid models with language encoded into the state representation. Full training curves for MiniGrid tasks only are located in Figure~\ref{fig:main_results_langstate}, IQM summary statistics for all environments (with only MiniGrid environments changed) are located in Figure~\ref{fig:iqm_langstate}, and Table~\ref{tab:full_numeric_table_langstate} contains updated raw performance numbers. As discussed in the main text, recall that MiniHack models all already encode language into the state representation.
We make the following observations about how these experiments differ from those in the main text:

\paragraph{Incorporating language into the feature space improves performance across all models.} However, just incorporating language in the feature space is insufficient for competitive performance alone: while the IMPALA baseline is able to make more progress in simpler environments (e.g.\ now nearly solving KeyCorridorS3R3), it still lags behind and is unable to solve the harder environments. This clearly illustrates that in order to reap the benefits of language, it is important to use it in conjunction with exploration, not just as a feature.

\paragraph{L-AMIGo continues to outperform AMIGo, though the differences are smaller.} While the tables show that both L-AMIGo and AMIGo can solve each task, reaching roughly similar final performance, they differ in sample efficiency as well as training stability (Figure~\ref{fig:main_results_langstate}, as well as the smaller error bars of L-AMIGo in Figure~\ref{fig:iqm_langstate}). Similar to the results in the main text, the full training curves in Figure~\ref{fig:main_results_langstate} show that while L-AMIGo and AMIGo reach similar asymptotic performance, L-AMIGo learns quicker or more stably. This happens quite clearly in KeyCorridorS5R3 and the Maze environments.

\paragraph{L-NovelD and NovelD performance on MiniGrid are similar.} This is the case except on a few tasks: KeyCorridorS4R3, and KeyCorridorS5R3 where NovelD is still unable to learn. Note that these NovelD results are somewhat unsurprising given the L-NovelD ablations and discussion in Appendix~\ref{app:lnoveld_ablations}; as we discuss there, L-NovelD is simply an adaptation of NovelD that enables a more precise tradeoff between language and state-based novelty. Similarly to how L-NovelD did not significantly outperform NovelD with language in the RND representation in Appendix~\ref{app:lnoveld_ablations}, L-NovelD does not outperform NovelD here, showing that for MiniGrid it is sufficient to naively combine the language and state representations. However, the MiniHack results in the main text indicate that there are settings where it is beneficial to have the separate L-NovelD term, and insufficient to solely encode language into the state representation.

To conclude, while adding language to the state representation results in smaller and more subtle performance differences in MiniGrid environments, the main conclusion (that language variants outperform their non-linguistic baselines) remains unchanged. As the last row of Table~\ref{tab:full_numeric_table_langstate} shows, overall, L-AMIGo outperforms AMIGo by 23\% (.16 absolute), and L-NovelD outperforms NovelD by 46\% (.24 absolute) across environments; both differences remain statistically significant.

\begin{table}[ht]
    \small
    \centering
    \caption{\textbf{IQM numbers for MiniGrid with language states.} Table elements same as Table~\ref{tab:full_numeric_table}.}
    \vspace{1em}
    \begin{tabular}{lrrrr}
    \toprule
    Environment & AMIGo & L-AMIGo & NovelD & L-NovelD \\
  \midrule
KeyCorridorS3R3 &	0.88 (0.88, 0.89) &	0.88 (0.88, 0.89) &	\textbf{0.89 (0.88, 0.89)} &	\textbf{0.89 (0.88, 0.89)} \\
KeyCorridorS4R3 &	0.88 (0.79, 0.90) &	0.89 (0.88, 0.89) &	0.89 (0.85, 0.90) &	\textbf{0.90 (0.89, 0.91)} \\
KeyCorridorS5R3 &	0.92 (0.90, 0.93) &	\textbf{0.93 (0.92, 0.93)} &	0.01 (0.00, 0.07) &	0.89 (0.07, 0.93) \\
ObstructedMaze\_1Dl	& 0.91 (0.41, 0.93) &	0.92 (0.90, 0.93) &	\textbf{0.93 (0.91, 0.93)} &	\textbf{0.93 (0.92, 0.93)} \\
ObstructedMaze\_2Dlhb	& 0.81 (0.34, 0.87) &	0.86 (0.83, 0.87) &	\textbf{0.89 (0.85, 0.91)} &	0.87 (0.74, 0.88) \\
ObstructedMaze\_1Q	& 0.92 (0.90, 0.93) &	0.91 (0.88, 0.92) &	0.92 (0.91, 0.94) &	\textbf{0.93 (0.93, 0.94)} \\
\midrule
\textbf{MiniGrid}	& 0.88 (0.79, 0.90) &	\textbf{0.90 (0.89, 0.90)} &	0.75 (0.74, 0.76) &	\textbf{0.90 (0.76, 0.91)} \\
\textbf{Overall} (w/ MiniHack) &	0.69 (0.65, 0.73) &	\textbf{0.85 (0.77, 0.90)} &	0.52 (0.47, 0.57) &	\textbf{0.76 (0.65, 0.83)} \\
    \bottomrule
    \end{tabular}
    \label{tab:full_numeric_table_langstate}
\end{table}

\section{Ablations}
\label{app:ablations}

\begin{figure}[t]
    \begin{minipage}[t]{0.48\textwidth}
        \centering
        \includegraphics[width=\linewidth]{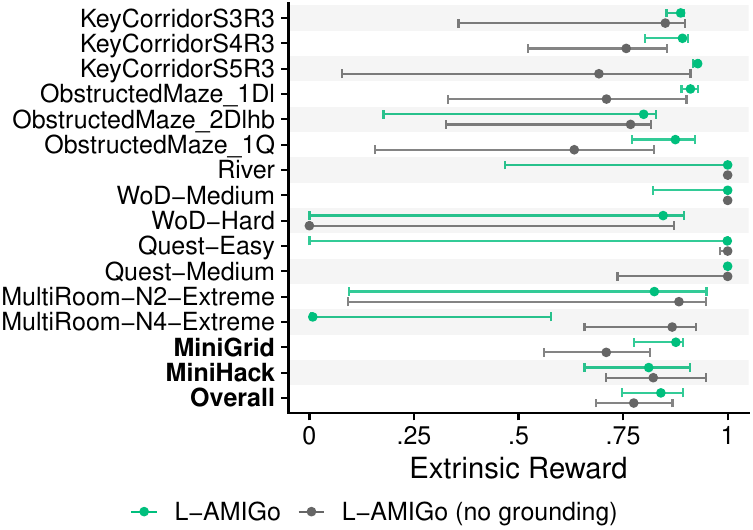}
        \caption{\textbf{L-AMIGo grounding network ablation}. IQM of L-AMIGo with and without grounding network across environments. Plot elements same as Figure~\ref{fig:iqm}.}
        \label{fig:lamigo_grounding}
    \end{minipage}
    \hfill
    \begin{minipage}[t]{0.48\textwidth}
        \centering
        \includegraphics[width=\linewidth]{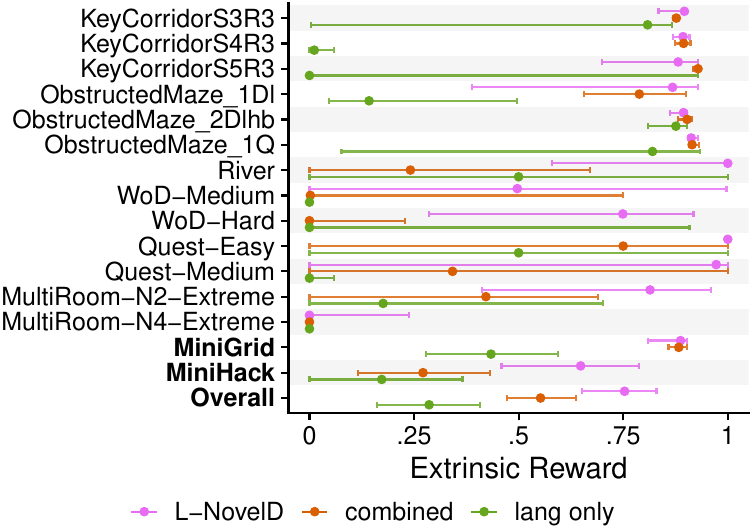}
        \caption{\textbf{L-NovelD ablations}. IQMs for Full L-NovelD, NovelD with combined state/language input, and language-only L-NovelD across environments. Plot elements same as Figure~\ref{fig:iqm}.}
        \label{fig:lnoveld_ablations}
    \end{minipage}
    \vspace{-1em}
\end{figure}

\subsection{L-AMIGo grounding network} 
\label{app:lamigo_ablations}

Figure~\ref{fig:lamigo_grounding} shows performance of L-AMIGo without the grounding network, where the policy network directly produces a distribution over goals without first predicting goal achievability, aggregated across domains. Overall, L-AMIGo without the grounding network performs reasonably well, matching full L-AMIGo performance on MiniHack. On MiniGrid, we see a modest difference in aggregate performance, though importantly, we also see greatly increased training stability with the grounding network on individual environments. Specifically, the confidence intervals are much larger for L-AMIGo without the grounding network on \textbf{KeyCorridorS\{3,4,5\}R3} and \textbf{ObstructedMaze\_\{1Dl,1Q\}}.

We hypothesize that the difference between MiniGrid and MiniHack tasks is because MiniGrid goals differ more between episodes: for example, since the colors of the doors are randomly shuffled, not all environments have red doors, so it is helpful to explicitly predict such features of the environment with a grounding network. In contrast, MiniHack envs have a more consistent set of goals (e.g.\ every WoD seed has a wand, a minotaur, etc), and so the grounding network is less necessary in this case.

\subsection{L-NovelD components}
\label{app:lnoveld_ablations}

To examine the relative performance of each component of L-NovelD, we run ablation experiments by (1) using the language-based reward \emph{only}, or (2) combining the language and state embedding into a single representation. As discussed in Appendix~\ref{app:naive_reward}. using the language-based reward only is equivalent to the naive message baseline (and thus prior work like \cite{goyal2019using,harrison2018guiding,mirchandani2021ella}) with an RND-determined novelty-based decay. 

Results are in Figure~\ref{fig:lnoveld_ablations}. They show that using the language reward alone results in uniformly worse performance across environments, suggesting that a naive message-based reward, even with novelty decay, underperforms and it is important to provide a simpler navigation-based bonus to encourage exploration.\footnote{Note also that message-only NovelD underperforms L-AMIGo as well.} Additionally, while combining the state and language into a single embedding works well for MiniGrid, it does not work as well for MiniHack tasks, suggesting that the additional flexibility afforded by the separate L-NovelD term can be helpful in many settings. In principle, it should be possible to tune the state and language embedding sizes of combined NovelD to see comparable performance to L-NovelD, but the point of L-NovelD is to clarify the contributions made by both the language and state and make it easier to trade-off between the two.

\section{Full description of tasks and language}
\label{app:full_tasks}

Here we describe each task in MiniGrid and MiniHack in detail, and enumerate the list of messages available in each task.
Examples of all tasks explored in this work are located in Figure~\ref{fig:full_tasks}.

\subsection{MiniGrid}
\label{app:full_minigrid_tasks}

\begin{figure*}
    \centering
    \includegraphics[width=\linewidth]{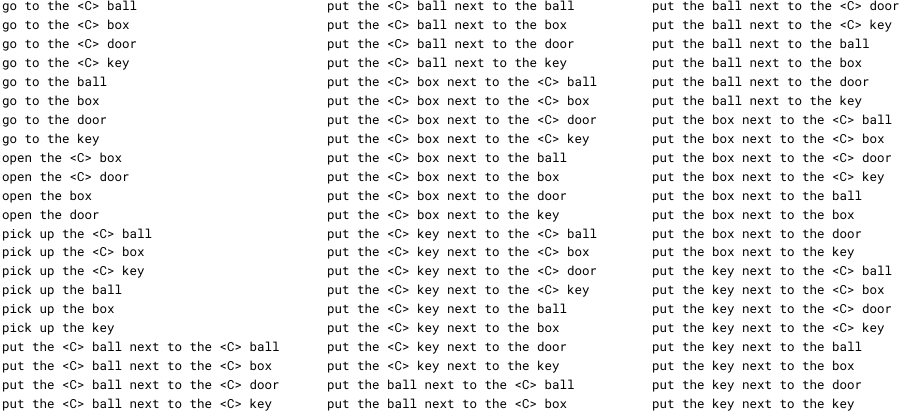}
    \caption{\textbf{Full list of possible MiniGrid messages}. Messages are synthesized with the BabyAI \cite{chevalierboisvert2019babyai} grammar, then divided into 66 templates. \texttt{<C>} is one of 6 possible colors: \emph{grey}, \emph{green}, \emph{blue}, \emph{red}, \emph{purple}, \emph{yellow}.}
    \label{fig:full_minigrid_messages}
\end{figure*}

\paragraph{Language.} As described in Section~\ref{sec:environments}, the set of possible descriptions in the BabyAI language 652 involves \emph{goto, open, pickup}, and \emph{putnext} commands which can be applied to boxes, doors, and balls, optionally qualified by color. These messages can be grouped into 66 message ``templates'' where specific colors are replaced with a placeholder \texttt{<C>}, as shown in Figure~\ref{fig:full_minigrid_messages}. Not all messages are needed for success on MiniGrid tasks, nor achieved by expert policies during training. We explain which messages are needed and/or encountered for each task below.

\paragraph{KeyCorridorS\{3,4,5\}R3.} In these tasks (Figure~\ref{fig:full_tasks}a--c), the agent is tasked with picking up a ball behind a locked door. To do so, it must first find and pick up the key which is hidden in (possibly several) nested rooms, return to the locked door, unlock the door, place the key down, and pick up the ball.
The agent start location, object colors, and room/door positions are randomized across seeds.

A typical policy trained on each task will encounter anywhere from 92 (KeyCorridorS3R3) to 141 messages (KeyCorridorS\{4,5\}R3) throughout training. Regardless of the environment size, a successful trajectory requires encountering anywhere from 8--12 messages: \emph{go to the door} (up to 3x), \emph{open the door} (up to 3x), \emph{go to the key}, \emph{pick up the key}, \emph{go to the} [locked] \emph{door}, \emph{open the} [locked] \emph{door}, \emph{go to the ball}, \emph{pick up the ball}.

\paragraph{ObstructedMaze-\{1Dl,2Dlhb,1Q\}.} In these tasks (Figure~\ref{fig:full_tasks}d--f), the agent is also tasked with picking up a ball behind a locked door. In the easier 1Dl task, the key is in the open; in the harder tasks, the keys are hidden in boxes which must be opened, and the doors are blocked by balls which must be moved to access them.

A typical policy trained on each task will encounter anywhere from 66 messages (ObstructedMaze-1Dl) to 244 messages (ObstructedMaze-1Q) throughout training. Of these,
a successful trajectory requires anywhere from 6 messages in ObstructedMaze-1Dl (\emph{go to the key}, \emph{pick up the key}, \emph{go to the door}, \emph{open the door}, \emph{go to the ball}, \emph{pick up the ball}) to 11 messages in ObstructedMaze-1Q (\emph{go to the door}, \emph{open the door}, \emph{go to the box}, \emph{open the box}, \emph{pick up the key}, \emph{go to the ball}, \emph{pick up the ball}, \emph{go to the door}, \emph{open the door}, \emph{go to the ball}, \emph{pick up the ball}).

\subsection{MiniHack}
\label{app:full_minihack_tasks}

\paragraph{Language.} As discussed previously, it is difficult to enumerate all messages in MiniHack, though we describe the messages encountered for each environment below, with a raw dump in Appendix~\ref{app:full_minihack_messages}. We perform some preprocessing on the messages to prevent unbounded growth: we replace all numbers (e.g.\ \emph{3 flint stones}, \emph{2 flint stones}, etc) with a single variable \emph{N}; we fix wands encountered in the environment to be a single Wand of Death (instead of having a wand of over 30 different types); items that can be arbitrarily named with random strings (e.g.\ \emph{a scroll labeled zelgo mer}; \emph{a dog named Hachi} have their names removed; finally, we cap the number of unique messages for each environment at 100.

\begin{figure*}
    \centering
    \includegraphics[width=\linewidth]{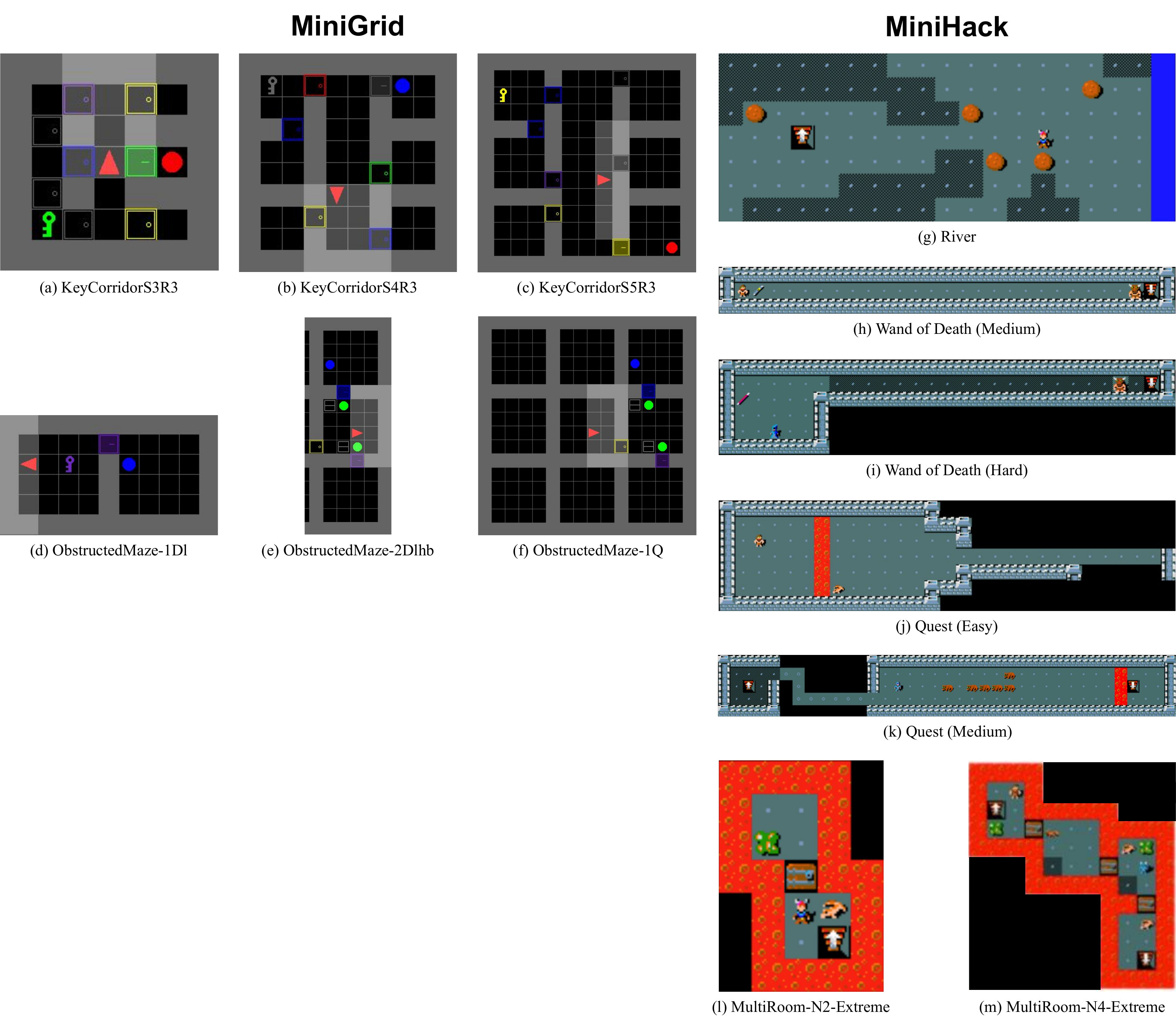}
    \caption{\textbf{Examples of all tasks evaluated in this work.} (a--f) MiniGrid tasks; (g--m) MiniHack tasks.}
    \label{fig:full_tasks}
\end{figure*}

\paragraph{River.} In this task (Figure~\ref{fig:full_tasks}g), the agent must cross the river located to the right on the environment, which can only be done by planning and pushing at least two boulders into the river in a row (to form a bridge over the river). A typical policy will encounter 14 distinct messages during training (Appendix~\ref{app:full_minihack_messages}.1), of which 7--8 messages (3--4 unique) are seen during training (\emph{with great effort you move the boulder}, \emph{you push the boulder into the water.}, \emph{now you can cross it!}; these messages must be repeated twice).

\paragraph{Wand of Death (\{Medium,Hard\}).}
In these tasks (Figure~\ref{fig:full_tasks}i--j; also described in the main text), the agent must learn to use a \emph{wand of death}, which can zap and kill enemies. This involves a complex sequence of actions: the agent must find the wand, pick it up, choose to \emph{zap} an item, select the wand in the inventory, and finally choose the direction to zap (towards the minotaur which is pursuing the player). It must then proceed past the minotaur to the goal to receive reward. Taking these actions out of order (e.g.\ trying to \emph{zap} something with nothing in the inventory, or selecting something other than the wand) has no effect.

In the Medium environment, the agent is placed in a narrow corridor with the wand somewhere in the corridor, and the minotaur is asleep (which gives the agent more time to explore). In the Hard environment, the agent is placed in a larger room where it must first find the wand, with the added challenge that the minotaur is awake and pursues the player, leading to death if the minotaur ever touches the player.

In both Medium and Hard environments, a policy encounters around 60 messages (Appendix~\ref{app:full_minihack_messages}.2), of which 7 messages (7 unique) are typically necessary to complete the task (\emph{you see here a wand}, \emph{f - a wand}, \emph{what do you want to zap?}, \emph{in what direction?} \emph{you klil the minotaur!}, \emph{welcome to experience level 2}, \emph{you see here a minotaur corpse}).

\paragraph{Quest (\{Easy,Medium\}).} These tasks (Figure~\ref{fig:full_tasks}j--k) require learning to use a \emph{Wand of Cold} to navigate over a river of lava while simultaneously fighting monsters. The agent spawns with a Wand of Cold in its inventory, and must learn to zap the Wand of Cold at the lava, which freezes it and forms a bridge over the lava.

In the Easy environment, the agent must first cross the lava river, then survive fights with one or two monsters to the staircase at the end of the hall. In the Medium environment, the agent must first fight several monsters \emph{before} crossing the lava river. There is an additional challenge: note the narrow 
corridor before the main room in the Medium environment. If the agent runs into the room and tries to fight the monsters all at once, it quickly will become overwhelmed and die; to successfully kill all monsters, the agent must learn to use the narrow corridor as a ``bottleneck'', first baiting then leading the monsters into the corridor,  until all are defeated. It can then use the Wand of Cold to cross the lava bridge to the staircase beyond.

In both Easy and Medium environments, a typical policy encounters the maximum of 100 messages (Appendix~\ref{app:full_minihack_messages}.3). 
Very few messages, besides \emph{what do you want to zap? / in what direction ? / the lava cools and solidifies} are \emph{required} to complete an episode, since there is a large variety of monsters faced and uncertainty in their behaviors, which can trigger additional messages. However, highly efficient expert play typically encounters 8--12 messages (7--10 unique) in the Easy environment and 30--40 messages (8--12 unique) in the Medium environment: besides freezing the lava to cross the environment, the rest of the messages are combat related (e.g.\ \emph{you kill the enemy!}, \emph{an enemy corpse}, additional zapping commands, etc).

\paragraph{MultiRoom-N\{2,4\}-Extreme.} These tasks (Figure~\ref{fig:full_tasks}l--m) are ported from the MiniGrid MultiRoom tasks, but with significant additional challenges. The ultimate task is to reach the last room in a sequence of interconnected rooms, but in the Extreme versions of this task, the walls are replaced with lava (which result in instant death if touched), there are monsters in each room (which must be fought), and additionally the doors are locked and must be ``kicked'' open (which requires that select the kick action, then kick in the direction of the door).

In both environments, a typical policy encounters the maximum of 100 messages (Appendix~\ref{app:full_minihack_messages}.4). In a successful trajectory, an agent encounters 20--30 (12--14 unique) messages in MultiRoom-N2 and 60--70 messages (16--18 unique) in MultiRoom-N4. These messages are mostly combat-related (\emph{the enemy hits!}, \emph{you hit the enemy!}, \emph{you kill the enemy}, with repeated messages relating to kicking down doors (\emph{in what direction?}, \emph{as you kick the door, it crashes open}, \emph{WHAAAAM!}, sometimes repeated up to 3 times per door, of which there are 1--3 doors). 

\section{Raw MiniHack Messages}
\label{app:full_minihack_messages}

Messages are located on the next page. Each shows the list of messages encountered by a single agent trained on an environment in the corresponding categories. Separate training runs will encounter different messages.

\begin{figure*}[t]
    \centering
    \includegraphics[width=\linewidth]{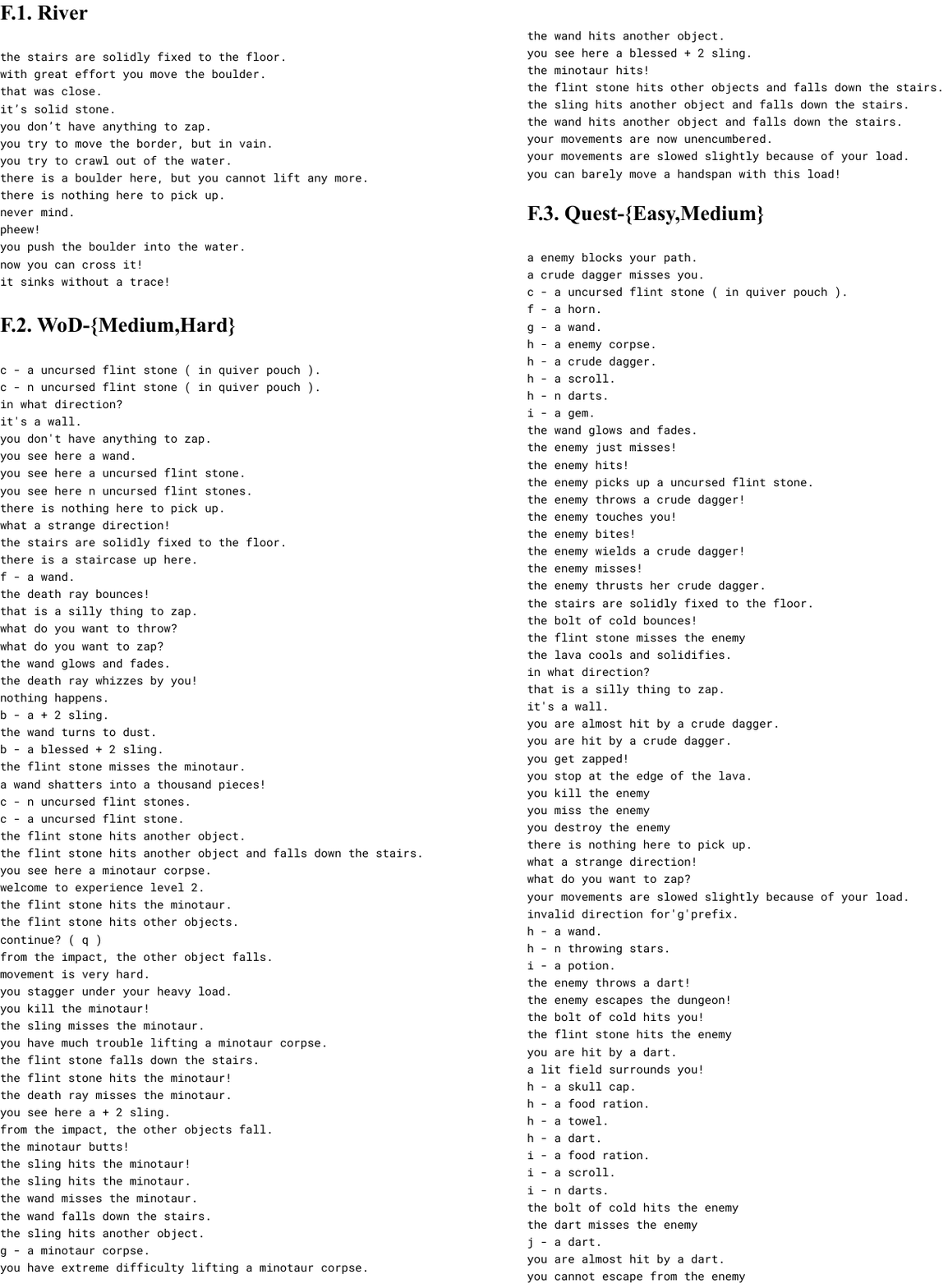}
\end{figure*}

\begin{figure*}[t]
    \centering
    \includegraphics[width=\linewidth]{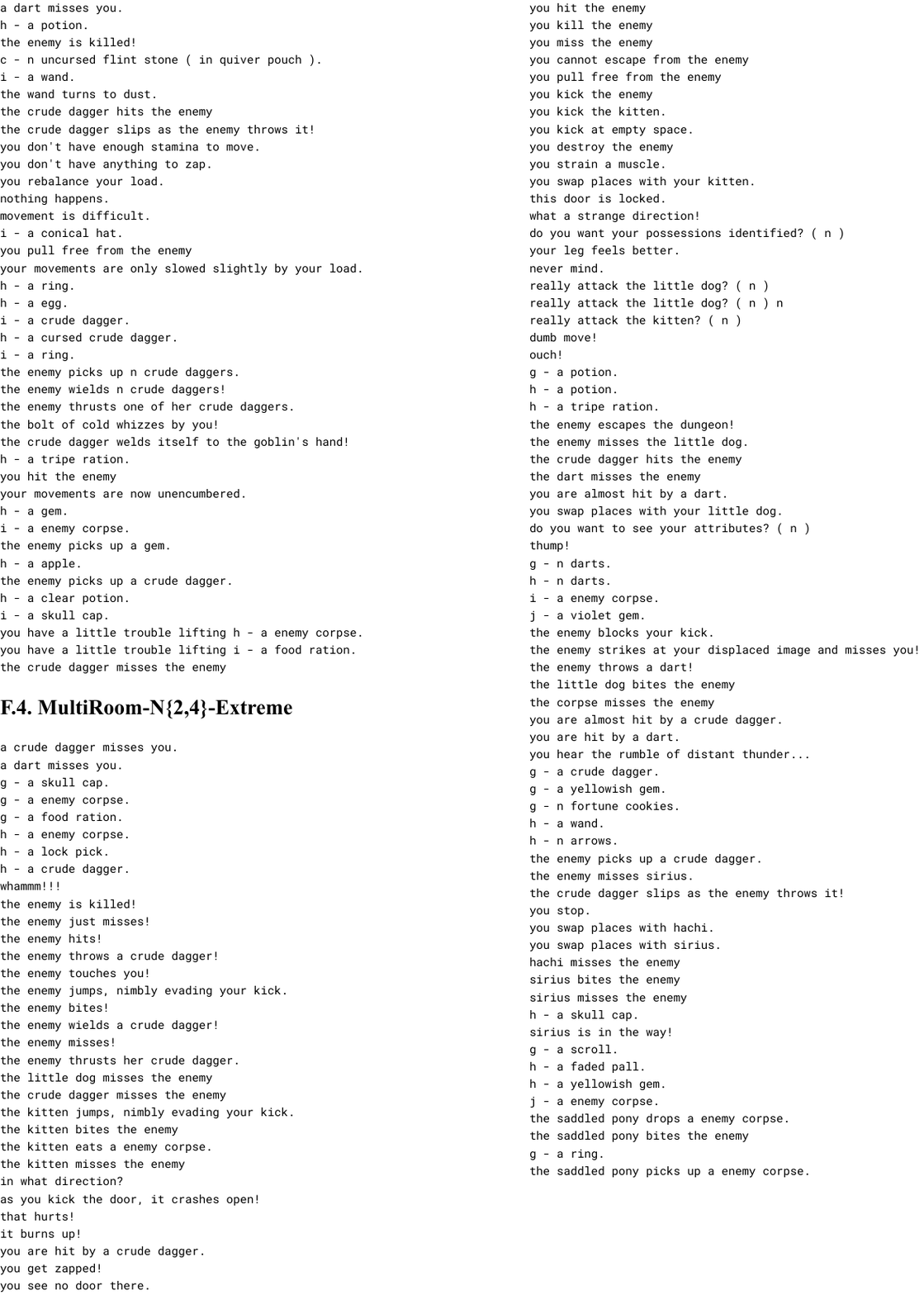}
\end{figure*}

\end{document}